\documentclass[11pt]{article}

\usepackage{acl}

\usepackage{times}
\usepackage{latexsym}

\usepackage[T1]{fontenc}

\usepackage[utf8]{inputenc}

\usepackage{microtype}

\usepackage{inconsolata}

\usepackage{graphicx}
\usepackage{subcaption}
\usepackage{booktabs}
\usepackage{multirow}
\usepackage{amsmath} 
\usepackage{amssymb}
\usepackage{placeins}
\title{Inter-3D VQA: A Roadside Multimodal Benchmark for 3D Spatiotemporally Grounded Visual Question Answering}

\author{
  Shaozu Ding \quad
  Linan Song \quad
  Dajiang Suo\thanks{Corresponding author.} \\
  The Polytechnic School, Arizona State University \\
  \texttt{\{sding32, lsong124, dsuo2\}@asu.edu}
}

\begin{document}
\maketitle

\begin{abstract}
Recent advances in visual question answering (VQA) and multimodal large language models (MLLMs) have enabled natural-language reasoning over traffic scenes. However, existing benchmarks are largely built from ego-vehicle views or 2D roadside videos, limiting their ability to evaluate 3D-grounded reasoning over real-world distances, trajectories, infrastructure topology, and safety-critical interactions. We introduce Inter-3D VQA, a large-scale roadside multimodal benchmark for 3D spatiotemporally grounded VQA at intersections. Built from synchronized point clouds and multi-view images, Inter-3D VQA contains 407K QA pairs covering lane-level positions, object relationships, motion patterns, and near-miss-oriented interaction reasoning. We further propose Inter-Geo, an MLLM baseline that integrates object- and scene-level aligned LiDAR representations, and Inter-Metrics, a unified evaluation framework for textual consistency, numerical accuracy, and semantic correctness. Experiments show that Inter-Geo outperforms image-based VLMs, especially on grounded spatial and temporal reasoning tasks. Our benchmark and codes are available at \url{https://github.com/ASU-Suo-Lab/Inter-3D-VQA}.
\end{abstract}

\section{Introduction}
Recent advances in multimodal large language models (MLLMs)~\cite{llava} have shifted traffic scene understanding from task-specific perception~\cite{PointPillars} toward language-guided reasoning~\cite{NuScenesQA}. Visual question answering (VQA)~\cite{VQA} enables users and downstream systems to query complex traffic scenes through natural language, supporting driving-oriented reasoning~\cite{Talk2Car,DriveGPT4} such as identifying relevant road users and explaining potential risks. Beyond onboard systems, VQA also offers a promising interface for infrastructure-side traffic monitoring and control, where operators may ask about crosswalk occupancy, lane blockage, pedestrian distance to conflict points, or safety-critical interactions.
\begin{figure}[t]
    \centering
    \includegraphics[width=\linewidth]{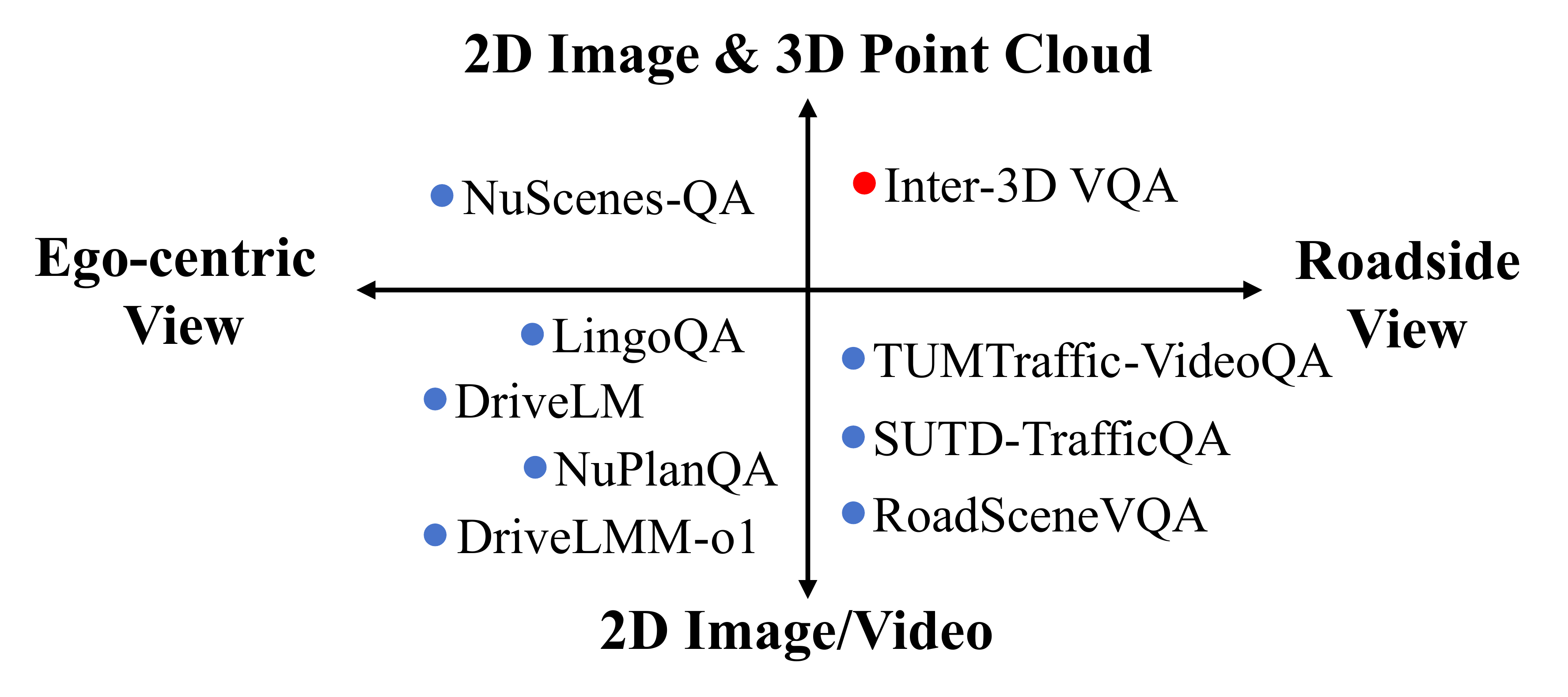}
    \caption{Inter-3D VQA is the first roadside multimodal benchmark for 3D grounded VQA.}
    \label{fig:innovation}
\end{figure}

However, existing traffic VQA benchmarks are still limited for 3D-grounded reasoning at intersection scale. As shown in Fig.~\ref{fig:innovation}, ego-centric benchmarks~\cite{drivelm,nuplanqa} rely on onboard sensors and suffer from occlusion and local viewpoints, while roadside benchmarks~\cite{TUMTrafficVideoQA,RoadSceneVQA} offer infrastructure-side perspectives but mainly use single-view images or 2D videos. As a result, they lack explicit 3D grounding for evaluating models’ understanding of real-world distances, trajectories, lane-level positions, and spatial relations among road users and infrastructure.

\begin{figure*}[t]
    \centering
    \includegraphics[width=\textwidth]{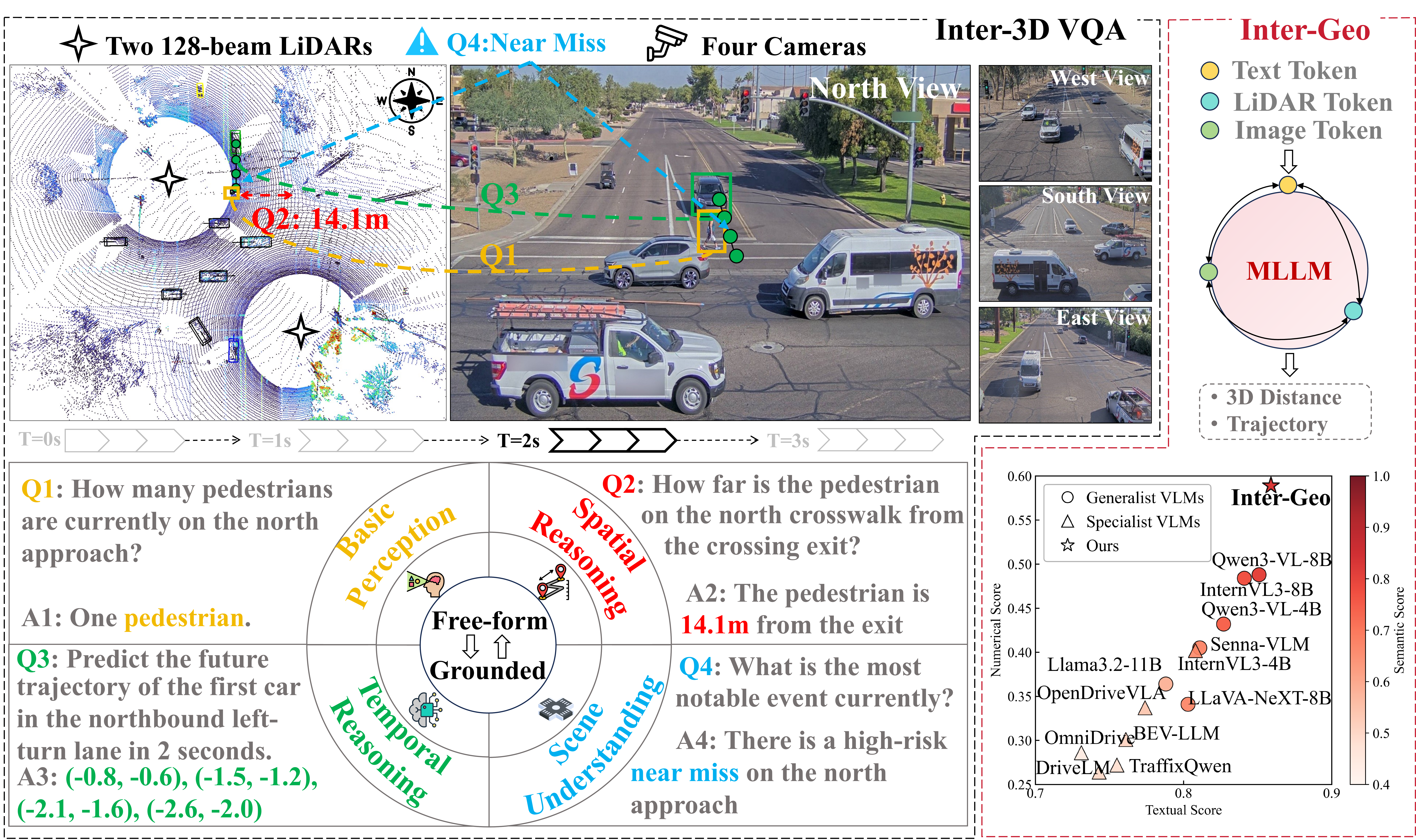}
    \caption{\textbf{Inter-3D VQA} is a large-scale roadside multimodal benchmark, aligning multi-view images with global point clouds to support diverse tasks from 3D perspective. Our MLLM \textbf{Inter-Geo} leverages object- and scene-level aligned LiDAR representations, achieving state-of-the-art performance on 3D grounded reasoning tasks.}
    \label{fig:overview}
\end{figure*}

This limitation is critical for traffic monitoring and safety applications. As illustrated in Fig.~\ref{fig:overview}, a system may need to localize a pedestrian in 3D space (Q1), determine whether the pedestrian remains within a crosswalk (Q2), and predict the trajectory of an approaching left-turning vehicle (Q3) to assess whether their future paths could result in a near-miss event (Q4), where road users come dangerously close without collision and signal potential safety risks before a crash occurs. Such reasoning requires spatiotemporal grounding in physical 3D space beyond image-level semantics. Due to the lack of such data and benchmarks, the community still lacks a systematic way to develop and evaluate MLLMs for bird’s-eye-view traffic understanding and 3D grounded VQA at intersections.

To address these challenges, we introduce Inter-3D VQA, a large-scale roadside multimodal benchmark built on synchronized global point clouds and multi-view images, as shown in Fig.~\ref{fig:overview}. It explicitly integrates 3D spatial information and infrastructure topology into VQA tasks. We further propose Inter-Geo, an MLLM that aligns object- and scene-level LiDAR representations via Dual Querying Net (DQNet) and injects them into the decoder through a LiDAR Decoder Adapter (LDA). We also introduce Inter-Metrics, a unified evaluation framework for textual consistency, numerical accuracy, and semantic correctness.
The main contributions of this work are summarized as follows:
\begin{itemize}
    \item We present \textbf{Inter-3D VQA}, a large-scale multimodal dataset for intersection understanding. It contains 407K QA pairs built from multi-view images, global point clouds, and infrastructure topology, enabling 3D-aware reasoning in complex roadside scenes.
    \item We propose a comprehensive benchmark for 3D roadside VQA, covering two QA formats and four task categories. We further introduce \textbf{Inter-Metrics}, a unified evaluation protocol that jointly measures textual consistency, numerical accuracy, and semantic correctness.
    \item We propose \textbf{Inter-Geo}, a multimodal large language model that incorporates object- and scene-level aligned LiDAR representations. Extensive experiments demonstrate its advantages over image-based VLMs, especially in 3D spatiotemporal reasoning.
\end{itemize}

\begin{table*}[t]
\centering
\small
\renewcommand{\arraystretch}{1.2}
\begin{tabular}{lccccccc}
\hline
\textbf{Dataset} & \textbf{Domain} & \textbf{Main Tasks} & \textbf{QA Generation} & \textbf{QA Type} & \textbf{Modality} & \textbf{QAs} \\
\hline
NuScenes-QA~\citeyearpar{NuScenesQA}  & Driving & 3D VQA & Temp. & Free-form & Image + PC & 460k \\
LingoQA~\citeyearpar{lingoqa} & Driving & VQA & Man. + LLM & Free-form & Image & 419k \\
DriveLM~\citeyearpar{drivelm} & Driving & GVQA + T.P. & Man. + Temp. & Grounded & Image & 443k \\
DriveLMM-o1~\citeyearpar{DriveLMM}  & Driving & M.C. & Man. + LLM & Grounded & Image & 18k \\
NuPlanQA~\citeyearpar{nuplanqa}  & Driving & M.C. & Man. + LLM & Free-form & Image & 1M \\
SURDS~\citeyearpar{SURDS} & Driving & VQA & Temp. + LLM & Free-form & Image & 50.3k \\
STSBench~\citeyearpar{STSBench}  & Driving & M.C. & Temp. & Free-form & Image  & 971 \\
Impromptu VLA~\citeyearpar{Impromptuvla}  & Driving & T.P. & Man. + LLM & Free-form & Image & 80k \\
AutoDrive-QA~\citeyearpar{AutoDriveQA} & Driving & M.C. & Man. + LLM & Grounded & Image & 15.4k \\
STRIDE-QA~\citeyearpar{STRIDEQA}  & Driving & VQA & Temp. & Grounded & Video & 16M \\
V2X-QA~\citeyearpar{V2X-QA}  & V2X & M.C. & Man. + LLM & Free-form & Image & 33k \\
City-3DQA~\citeyearpar{cityqa}  & City & 3D VQA & Temp. + LLM + Man. & Free-form & Point Cloud  & 450k \\
Open3D-VQA~\citeyearpar{open3dvqa} & City & 3D VQA & Temp. + LLM + Man. & Free-form & Image  & 73.3k \\
SUTD-TrafficQA~\citeyearpar{SUTDTrafficQA}  & Roadside & VQA & Man. & Free-form & Video  & 62k \\
TUMTraffic-VideoQA~\citeyearpar{TUMTrafficVideoQA}  & Roadside & VQA + M.C. & Temp. + LLM & Grounded & Video & 87.3k \\
RoadSceneVQA~\citeyearpar{RoadSceneVQA} & Roadside & VQA & Man. + LLM & Free-form & Image & 34.7k \\
\hline
Inter-3D VQA (ours) & Roadside & 3D VQA + T.P. & Temp. + LLM + Man. & Both & Image + PC & 407k  \\
\hline
\end{tabular}
\caption{Comprehensive comparison of representative VQA benchmarks in autonomous driving, urban, and roadside scenarios. Temp.: Template, Man.: Manual, M.C.: Multiple Choice, T.P.: Trajectory Prediction, PC: Point Cloud.}
\label{tab:vqa_benchmark_comparison}
\end{table*}

\section{Related Works}
Existing VQA research in transportation spans autonomous driving, urban scenes, and roadside understanding, with substantial differences in sensing perspectives, tasks, and data modalities. Table~\ref{tab:vqa_benchmark_comparison} summarizes representative VQA benchmarks across these scenarios in terms of task design, construction method, QA type, modality, and scale.
\subsection{Autonomous Driving VQA}
Existing autonomous driving VQA benchmarks are predominantly built on ego-centric views, focusing on driving-oriented reasoning. NuScenes-QA~\cite{NuScenesQA}, LingoQA~\cite{lingoqa} and DriveLM~\cite{drivelm} establish the task through multimodal QA, video-based QA, and graph-structured reasoning over perception, prediction, and planning. Recent benchmarks, including DriveLMM-o1~\cite{DriveLMM}, NuPlanQA~\cite{nuplanqa}, AutoDrive-QA~\cite{AutoDriveQA}, SURDS~\cite{SURDS}, STSBench~\cite{STSBench}, and STRIDE-QA~\cite{STRIDEQA}, further emphasize multi-view perception and spatio-temporal reasoning. However, these benchmarks remain constrained to vehicle-mounted perspectives, limiting their ability to evaluate global intersection reasoning and infrastructure-side traffic understanding.

\subsection{Roadside and Urban Outdoor VQA}
Compared with ego-centric benchmarks, roadside VQA focuses on fixed-view scene understanding, enabling better modeling of global traffic dynamics and multi-agent interactions. SUTD-TrafficQA~\cite{SUTDTrafficQA} formulates traffic event understanding as QA. TUMTraffic-VideoQA~\cite{TUMTrafficVideoQA} and RoadSceneVQA~\cite{RoadSceneVQA} extend this paradigm to roadside scenarios with video QA, grounding, and higher-level semantics. Meanwhile, City-3DQA~\cite{cityqa} and Open3D-VQA~\cite{open3dvqa} extend QA to urban environments from a drone-based perspective. However, existing benchmarks still rely on single-modality observations and lack explicit 3D geometric information, limiting their ability to support multimodal reasoning that integrates multi-directional views and 3D spatial structure.

\subsection{Vision-Language Models for Traffic Scene}
Existing traffic VLMs for scene understanding span camera-only, LiDAR-only, and multimodal paradigms. Camera-based models such as OmniDrive~\cite{OmniDrive} and OpenDriveVLA~\cite{OpenDriveVLA} rely on image inputs, while TraffiX-Qwen~\cite{TUMTrafficVideoQA} extends them to roadside scenarios. LiDAR-based and multimodal methods such as LiDAR-LLM~\cite{LiDARLLM}, NuScenes-QA~\cite{NuScenesQA}, and BEV-LLM~\cite{BEVLLM} incorporate LiDAR inputs but are designed for ego-centric settings. In contrast, roadside models remain camera-based, limiting their ability to capture global 3D structure and spatial relationships. Our Inter-Geo addresses this gap by combining multi-view images with global point clouds, enabling cross-view semantics and geometry-aware reasoning.

\begin{figure*}
    \centering
    \includegraphics[width=\textwidth]{pdf/qa_construction_pipeline_newest.pdf}
    \caption{Data construction pipeline of Inter-3D VQA. The pipeline consists of three stages: metadata curation from synchronized multi-view images, LiDAR point clouds, and HD maps, template-based QA generation with LLM-assisted expansion, and a multi-stage sanity check for quality control and data balancing.}
    \label{fig:qa_construction_pipeline}
\end{figure*}

\section{Inter-3D VQA Dataset}
\subsection{Dataset Construction}
As shown in Fig.~\ref{fig:qa_construction_pipeline}, inspired by prior VQA dataset construction pipelines~\cite{NuScenesQA,drivelm}, we leverage open-source roadside perception data~\cite{ding2026resolve} to develop an automated QA generation framework that combines template filling with AI-assisted annotation and validation. All data are de-identified by anonymizing faces and license plates using automatic detection, manual verification, and Gaussian blurring.

\textbf{Metadata Curation.} We utilize synchronized multi-view video and LiDAR point cloud streams, sampling keyframes at 2 Hz to construct image–point cloud pairs. After manual filtering, the raw data covers diverse environmental conditions, traffic patterns, and challenging perception scenarios. We then extract structured metadata from these multimodal inputs. Environmental semantics (e.g., traffic light state) are extracted from images using VLMs, while LiDAR data with annotated 3D bounding boxes provides object-level attributes and temporal associations. In addition, High-Definition maps are aligned with point clouds to extract infrastructure elements, enabling topology-aware reasoning. Finally, all modalities are unified into a structured per-frame representation that captures comprehensive information about the environment, objects, and infrastructure.

\textbf{QA Generation.} To improve dataset diversity, we adopt a template-based generation strategy that combines expert knowledge with Large Language Models (LLMs). We first collect representative traffic questions from domain experts and use a domain-adapted LLM to generate derivative questions. To ensure quality, the generated questions are manually curated and grouped into standardized templates with reserved metadata slots. Unlike conventional approaches, our templates encode precise 3D spatial and infrastructure topology information. Finally, QA pairs are instantiated by filling templates with structured metadata.

\textbf{Sanity Check.} We employ a multi-stage quality control pipeline to ensure the accuracy and diversity of QA pairs, with human intervention only at key stages. QA pairs first undergo consistency checks to remove instances violating common sense or traffic rules, followed by ambiguity filtering to discard questions with multiple valid answers or unclear semantics. LLMs then generate semantically equivalent variants to enhance linguistic diversity. A sampled manual review corrects low-quality samples, and finally template- and instance-level sampling are applied to balance the distribution across task categories.

\subsection{Object Referring}
Existing roadside VQA datasets typically associate objects with language through structured grounding or free-form descriptions, both aiming to achieve unambiguous object references. Grounded representations, as used in TUMTraffic-VideoQA~\cite{TUMTrafficVideoQA}, support object localization and temporal correspondence. However, they rely on single-view observations from 2D images, which are insufficient to support spatial reasoning with accurate 3D information. Free-form descriptions, as used in RoadSceneVQA~\cite{RoadSceneVQA}, offer greater flexibility but often introduce ambiguity when multiple similar objects are present, leading to degraded model performance.

\begin{figure}[t]
    \centering
    \includegraphics[width=\linewidth]{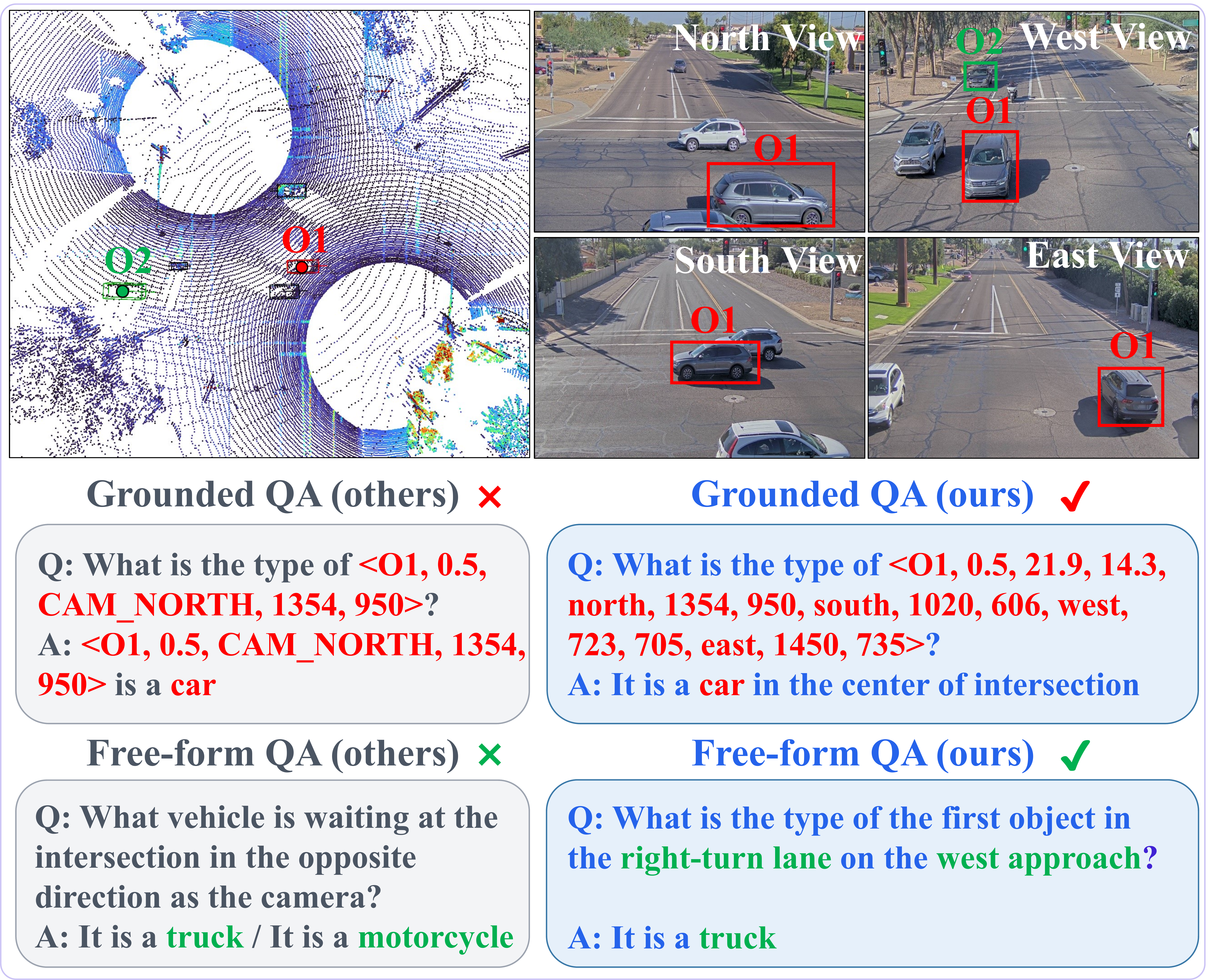}
    \caption{Object referring mechanisms in Inter-3D VQA. Grounded QA provides spatiotemporal object references for cross-view and 3D grounding, while free-form QA uses topology-aware language descriptions to reduce ambiguity through infrastructure constraints.}
    \label{fig:qa_types}
\end{figure}

To address these limitations, Inter-3D VQA extends prior object referring schemes~\cite{nuplanqa} into two complementary QA formats as shown in Fig.~\ref{fig:qa_types}. Free-form QA uses natural language descriptions enriched with infrastructure topology, enabling flexible reasoning over lanes, crosswalks, road structures, and traffic participants while reducing ambiguity among similar objects. Grounded QA provides explicit object references through a 3D-aware spatiotemporal tuple $<o, f_n, x_{3d}, y_{3d}, cam_{view}, x_{2d}, y_{2d}, \ldots>$, where $o$ denotes a unique object identifier, $f_n$ is the temporal offset, $(x_{3d}, y_{3d})$ represents the 3D object center, and $(cam_{view}, x_{2d}, y_{2d}, \ldots)$ encodes multi-view projections. This representation aligns point clouds, multi-view images, language references, and infrastructure topology, enabling precise cross-view association, 3D localization, temporal tracking, and topology-aware object disambiguation.

\subsection{Task Definition}
Existing roadside VQA benchmarks~\cite{TUMTrafficVideoQA,RoadSceneVQA} focus on image-based tasks, making it difficult to support precise 3D measurements (e.g., distance to a crosswalk), infrastructure-aware relations (e.g., lane-level topology), and interaction reasoning (e.g., potential hazards). To address this gap, Inter-3D VQA integrates multi-view images with global point clouds and organizes the benchmark into four task categories: Basic Perception, Spatial Reasoning, Temporal Reasoning, and Scene Understanding. These tasks cover capabilities from low-level perception to high-level reasoning in complex roadside scenarios. Inter-3D VQA contains 407K QA pairs, split into training and test sets at an 8:2 ratio with no scene overlap to prevent memorization bias. Although certain safety-critical events can be rare in a single-intersection case, our QA construction pipeline explicitly covers diverse traffic behaviors and safety-relevant interactions. Queue dynamics, multi-agent conflicts, and abnormal events account for 16.8\% of total QA pairs, while over 30\% involve speeding or lane-changing behaviors rather than regular passing. Detailed task definitions and dataset statistics are provided in the appendix.

\begin{figure*}[t]
    \centering 
    \includegraphics[width=\linewidth]{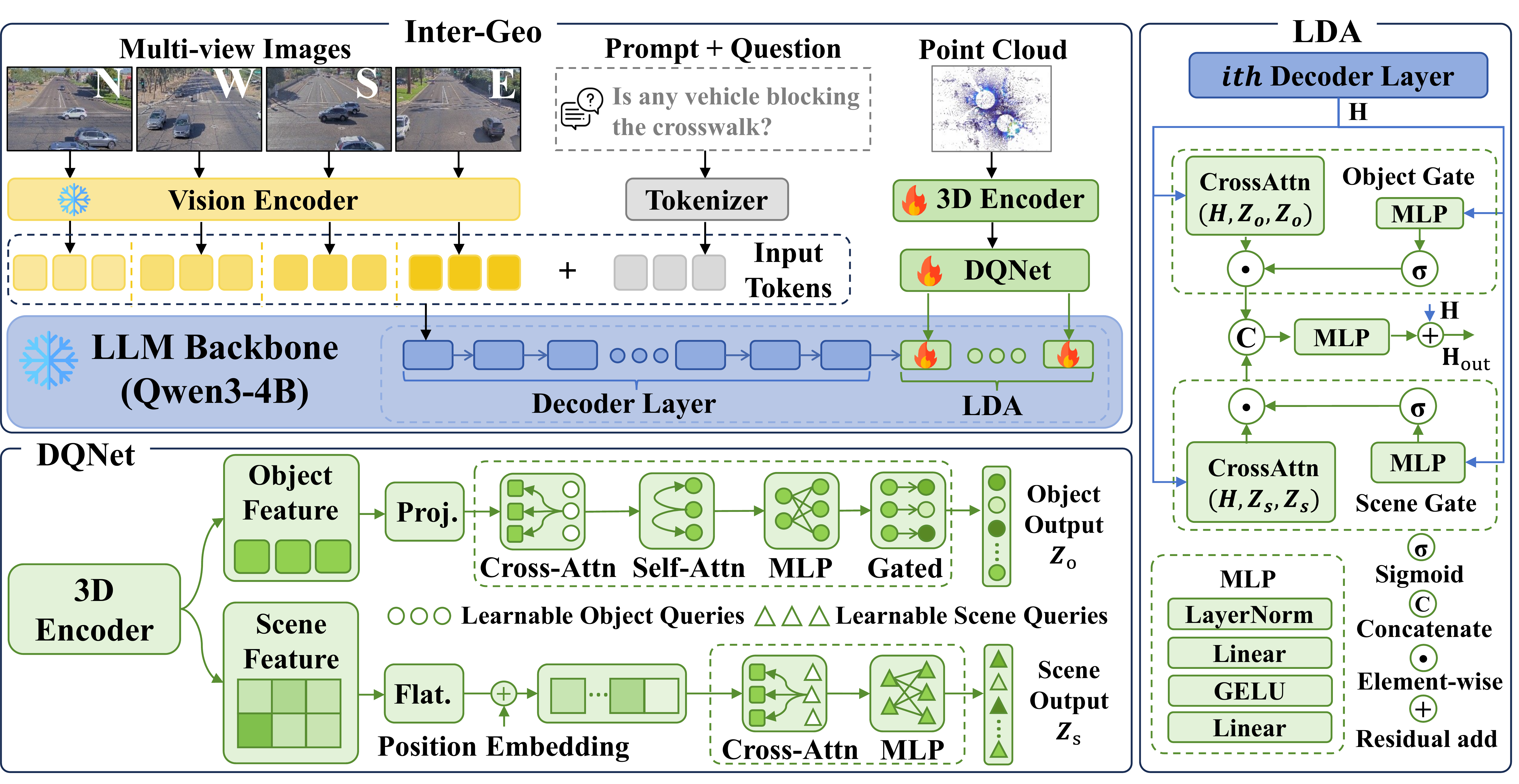}
    \caption{Architecture of Inter-Geo model, which integrates multi-view images, text, and global point clouds. Object- and scene-level features are extracted by 3D encoder, then transformed into dual aligned LiDAR representations via DQNet, and finally injected into the LLM decoder through LDA for geometry-aware reasoning.}
    \label{fig:Inter-Geo}
\end{figure*}

\subsection{Inter-Metrics} 
Existing roadside VQA benchmarks also exhibit limitations in evaluating 3D-grounded reasoning. RoadSceneVQA~\cite{RoadSceneVQA} relies on language generation metrics, which fail to capture fine-grained semantic differences (e.g., opposite traffic directions may yield similar scores). TUMTraffic-VideoQA advances roadside video understanding through spatio-temporal video grounding, but its grounding remains primarily in the image/video space, making it difficult to assess whether a model understands real-world 3D positions, distances, trajectories, and safety-critical interactions. For 3D-grounded traffic VQA, correct language is not enough. A model must also be numerically accurate in physical space and semantically correct with respect to traffic structure. We therefore propose Inter-Metrics, a unified framework that evaluates models across textual consistency, numerical accuracy, and semantic correctness.

\textbf{Textual Consistency.} For open-ended textual answers, we measure similarity between the predicted answer $\hat{y}$ and the reference $y$ using multiple metrics (BLEU-4~\citeyearpar{BLEU}, ROUGE-L~\citeyearpar{rouge}, BERTScore~\citeyearpar{BERTScore}, and SimCSE~\citeyearpar{SimCSE}), aggregated as:
\begin{equation}
\label{eq:text}
S_{\text{text}}(\hat{y}, y) =
\sum_{m \in \mathcal{M}} w_m \, s_m(\hat{y}, y)
\end{equation}
where $s_m$ denotes the score of metric $m$, and $w_m$ is its normalized weight satisfying
$\sum_{m\in\mathcal{M}} w_m = 1$.

\textbf{Numerical Accuracy.} We evaluate scalar fields $\mathcal{S}$ (e.g., count, speed, distances) using absolute error and vector fields $\mathcal{V}$ (e.g., 3D coordinates, trajectories) using Euclidean error:
\begin{equation}
\label{eq:numerical_sub}
e_i =
\begin{cases}
|\hat{v}_i - v_i|, & \text{if } i \in \mathcal{S}, \\
\|\hat{\mathbf{p}}_i - \mathbf{p}_i\|_2, & \text{if } i \in \mathcal{V},
\end{cases}
\end{equation}
To ensure comparability across heterogeneous quantities with different units and scales, errors are normalized by field-specific thresholds $\tau_i$ and mapped to a unified score in the range $[0,1]$:
\begin{equation}
S_{\text{num}} =
1 - \frac{1}{N} \sum_{i=1}^{N}
\min\left(\frac{e_i}{\tau_i}, 1\right)
\end{equation}
where $N$ is the number of evaluated fields. The same field-specific thresholds are applied consistently across all models to ensure a fair comparison.

\textbf{Semantic Correctness.} We evaluate semantic correctness by exact matching of structured semantic fields~\cite{VQA}, such as object categories, lane membership, movement direction, traffic states, and conflict labels, going beyond surface-level linguistic similarity:
\begin{equation}
S_{\text{sem}}(\hat{\mathbf{z}}, \mathbf{z}) =
\frac{1}{|\mathcal{F}|}
\sum_{f \in \mathcal{F}}
\mathbb{I}(\hat{z}_f = z_f)
\end{equation}
where $\mathbb{I}(\cdot)$ is the indicator function, and $\hat{z}_f$ and $z_f$ denote the predicted and ground-truth values of semantic field $f$, respectively.

\section{Methods}
\subsection{Overall Architecture}
As shown in Fig.~\ref{fig:Inter-Geo}, we propose Inter-Geo, a LiDAR-enhanced vision-language model built on Qwen3-VL-4B-Instruct. Following the Flamingo design~\cite{Flamingo}, we preserve the original camera–text pipeline and integrate LiDAR modality through decoder-level cross-attention, enabling geometry-aware reasoning without disrupting pretrained multimodal alignment. The model takes multi-view images $\mathcal{I}=\{I_N, I_W, I_S, I_E\}$ and text $\mathcal{T}$ as inputs. Images are encoded into visual embeddings $V = E_{\text{vis}}(\mathcal{I})$, which are inserted into the token sequence with text embeddings $T = E_{\text{tok}}(\operatorname{Tokenizer}(\mathcal{T}))$, forming the multimodal input $H^0 = [T; V]$ for the decoder.

In parallel, LiDAR input is processed by 3D encoder to extract object- and scene-level features $\{X_o, X_s\}$, which the Dual Querying Net converts into aligned LiDAR representations:
\begin{equation}
(Z_o, Z_s) = \operatorname{DQNet}(X_o, X_s)
\end{equation}

These representations are injected into selected decoder layers through LiDAR Decoder Adapter, where hidden states are updated via cross-attention:
\begin{equation}
H^{l+1} = \operatorname{LDA}^{l}(\operatorname{Dec}^{l}(H^l), Z_o, Z_s)
\end{equation}

\begin{table*}[ht]
\centering
\small
\renewcommand{\arraystretch}{1.1}
\begin{tabular}{llcccccc}
\toprule
\multirow{2}{*}{\textbf{Types}} & \multirow{2}{*}{\textbf{Models}} & \multicolumn{3}{c}{\textbf{Free-form QA}} & \multicolumn{3}{c}{\textbf{Grounded QA}} \\
\cmidrule(lr){3-5} \cmidrule(lr){6-8}
& & \textbf{$S_{\text{text}} \uparrow$} & \textbf{$S_{\text{num}} \uparrow$} & \textbf{$S_{\text{sem}} \uparrow$}
& \textbf{$S_{\text{text}} \uparrow$} & \textbf{$S_{\text{num}} \uparrow$} & \textbf{$S_{\text{sem}} \uparrow$} \\
\midrule

\multirow{6}{*}{Open-source Generalist VLMs}
& LLaVA-NeXT-8B~\citeyearpar{LLaVA-NeXT}  & 0.788 & 0.347 & 0.646 & 0.803 & 0.341 & 0.656 \\
& Llama3.2-11B~\citeyearpar{llama32}      & 0.770 & 0.289 & 0.583 & 0.788 & 0.364 & 0.576 \\
& Qwen3-VL-4B~\citeyearpar{Qwen3-VL}      & 0.806 & 0.397 & 0.709 & 0.827 & 0.432 & 0.736 \\
& Qwen3-VL-8B~\citeyearpar{Qwen3-VL}      & \underline{\textbf{0.821}} & \underline{0.457} & \underline{\textbf{0.749}} & \underline{0.851} & \underline{0.488} & \underline{0.802} \\
& InternVL3-4B~\citeyearpar{InternVL3}    & 0.794 & 0.363 & 0.656 & 0.811 & 0.405 & 0.666 \\
& InternVL3-8B~\citeyearpar{InternVL3}    & 0.814 & 0.435 & 0.724 & 0.841 & 0.484 & 0.768 \\

\midrule
\multirow{6}{*}{Specialized Driving VLMs}
& Senna-VLM~\citeyearpar{Senna}          & \underline{0.784} & 0.321 & \underline{0.628} & \underline{0.808} & \underline{0.402} & \underline{0.644} \\
& NuScenes-QA~\citeyearpar{NuScenesQA}     & 0.671 & 0.288 & 0.294 & 0.612 & 0.127 & 0.173 \\
& OmniDrive~\citeyearpar{OmniDrive}        & 0.752 & 0.336 & 0.560 & 0.731 & 0.286 & 0.440 \\
& OpenDriveVLA~\citeyearpar{OpenDriveVLA}  & 0.777 & \underline{0.421} & 0.624 & 0.774 & 0.337 & 0.534 \\
& DriveLM~\citeyearpar{drivelm}            & 0.766 & 0.381 & 0.582 & 0.755 & 0.272 & 0.481 \\
& BEV-LLM~\citeyearpar{BEVLLM}            & 0.772 & 0.376 & 0.613 & 0.761 & 0.301 & 0.507 \\

\midrule
\multirow{2}{*}{Specialized Roadside VLMs}
& TraffiX-Qwen~\citeyearpar{TUMTrafficVideoQA}         & 0.753 & 0.322 & 0.564 & 0.743 & 0.264 & 0.495 \\
& Inter-Geo (ours)      & \underline{0.820} & \underline{\textbf{0.482}} & \underline{0.748} & \underline{\textbf{0.859}} & \underline{\textbf{0.589}} & \underline{\textbf{0.828}}  \\

\bottomrule
\end{tabular}
\caption{Performance comparison of Inter-Geo with various VLMs on both free-form and grounded Inter-3D VQA.}
\label{tab:main_results}
\end{table*}

\subsection{3D Encoder}
Our framework is agnostic to 3D perception backbone. In practice, we adopt a LION-based backbone~\cite{LION} with a TransFusion detection head~\cite{TransFusion}, though any 3D detection model can be used. Given point clouds $\mathcal{P}$, the encoder extracts scene-level BEV features $X_s$ capturing global context, and object-level features $X_o$ describing individual instances. The subsequent DQNet operates solely on $X_s$ and $X_o$, enabling flexible integration with different backbones.

\subsection{Dual Querying Net}
The Dual Querying Net (DQNet) converts variable-length LiDAR features into fixed-size representations aligned with the language model latent space~\cite{BLIP-2}. It consists of an object branch and a scene branch, capturing instance-level details and global context.

In the object branch, we adopt DETR-inspired query aggregation~\cite{DETR3D}, where learnable queries $Q_a$ attend to projected object tokens $T_o$ and produce fixed object-aware representations refined by attention and MLP layers:
\begin{equation}
\hat{Q}_a = \operatorname{MLP}\!\left(
\operatorname{Attn}_{\text{self}}\!\left(
\operatorname{Attn}_{\text{cross}}(Q_a, T_o)
\right)
\right)
\end{equation}

A confidence gate is then applied to suppress unreliable or redundant representations:
\begin{equation}
Z_o = \sigma(W_c \hat{Q}_a)\odot \hat{Q}_a
\end{equation}
where $\sigma(\cdot)$ denotes the sigmoid function and $W_c$ is a learnable projection.

In the scene branch, BEV features are projected into the language model space as tokens $T_s$ and compressed using a Perceiver-style latent bottleneck~\cite{PerceiverIO}. Specifically, learnable queries $Q_s$ aggregate global context via cross-attention and are refined as:
\begin{equation}
Z_s = \operatorname{MLP}\!\left(
\operatorname{Attn}_{\text{cross}}(Q_s, T_s)
\right)
\end{equation}

\subsection{LiDAR Decoder Adapter}
The LiDAR Decoder Adapter (LDA) integrates LiDAR representations into the Qwen3-VL decoder~\cite{Flamingo}. Instead of appending LiDAR features as input tokens, LDA enables selective retrieval of geometric information while preserving the pretrained language–vision alignment. Given decoder states $H$, scene- and object-level information are retrieved via cross-attention:
\begin{equation}
\Delta_{\text{s/o}} = \operatorname{Attn}_{\text{cross}}(H, Z_{s/o})
\end{equation}

To regulate LiDAR contributions, we employ a token- and channel-wise gating mechanism~\cite{FiLM}, which adaptively scales LiDAR features based on the language hidden states:
\begin{equation}
G_{s/o} = \sigma(\operatorname{MLP_{s/o}(H)})
\end{equation}

The gated features are then fused and injected into the decoder states via a residual module:
\begin{equation}
H_{\text{out}} = H + 
\operatorname{MLP}_f \!\left(
[G_s \odot \Delta_{\text{s}};\ G_o \odot \Delta_{\text{o}}]
\right)
\end{equation}
where $[\cdot;\cdot]$ represents feature concatenation.

\section{Experiments}
\subsection{Experimental Settings}
For Inter-Geo, we adopt Qwen3-VL-4B-Instruct as the base model and perform LoRA-based fine-tuning using Llama-Factory~\cite{LlamaFactory}. We train the model using a learning rate of $1 \times 10^{-4}$ and cosine scheduling. It is trained for 3 epochs with a per-GPU batch size of 1 and gradient accumulation of 8. All experiments are conducted on four NVIDIA RTX PRO 6000 Blackwell GPUs. For a fair comparison, all open-source baselines are fine-tuned and evaluated using the same data split, prompts, grounded references, and evaluation protocol. Additional implementation details are provided in the appendix.

\subsection{Quantitative Results}
Table~\ref{tab:main_results} shows Inter-Geo achieves the best grounded QA performance among existing VLMs while remaining highly competitive on free-form QA. For free-form QA, Inter-Geo obtains the highest numerical score $S_{\text{num}}=0.482$, outperforming the strongest open-source baseline Qwen3-VL-8B by 5.5\%. Its textual and semantic scores are nearly identical to Qwen3-VL-8B, indicating that Inter-Geo preserves strong language understanding while improving geometric reasoning through aligned LiDAR representations. For grounded QA, Inter-Geo outperforms all baselines across all metrics. Compared to Qwen3-VL-8B, it improves numerical accuracy by 20.7\%. Compared with the top-performing driving-specific model, Senna-VLM, it improves $S_{\text{num}}$ and $S_{\text{sem}}$ by 46.5\% and 28.6\%. These gains suggest that the dual-branch LiDAR representations with decoder-level injection are effective for grounded QA that requires 3D localization and spatial reasoning. Although TraffiX-Qwen is roadside-specific, its image-dominant design relies on 2D cues for depth and distance reasoning, which can be unreliable under occlusion and cross-view ambiguity.

\begin{table}[t]
\small
\centering
\setlength{\tabcolsep}{4pt}
\begin{tabular}{c c c c c c c}
\toprule
\multirow{2}{*}{3D Enc.} 
& \multicolumn{2}{c}{DQNet} 
& \multirow{2}{*}{LDA} 
& \multirow{2}{*}{$S_{\text{text}} \uparrow$} 
& \multirow{2}{*}{$S_{\text{num}} \uparrow$} 
& \multirow{2}{*}{$S_{\text{sem}} \uparrow$} \\
\cmidrule(lr){2-3}
& Obj. & Scene & & & & \\
\midrule
\checkmark & \checkmark & \checkmark & \checkmark & 0.859 & 0.589 & 0.828 \\
           & \checkmark & \checkmark & \checkmark & 0.841 & 0.477 & 0.765 \\
\checkmark &            & \checkmark & \checkmark & 0.857 & 0.566 & 0.822 \\
\checkmark & \checkmark &            & \checkmark & 0.857 & 0.553 & 0.817 \\
\checkmark & \checkmark & \checkmark &            & 0.827 & 0.432 & 0.736 \\
\bottomrule
\end{tabular}
\caption{Contributions of LiDAR-related modules. Removing both DQNet branches is equivalent to removing LDA, as neither provides LiDAR information.}
\label{tab:lidar_modules}
\end{table}

\subsection{Ablation Studies}
\textbf{Effect of LiDAR-related Modules.}
Table~\ref{tab:lidar_modules} evaluates LiDAR-related modules on grounded Inter-3D VQA. Replacing the 3D encoder with a MLP encoder reduces $S_{\text{num}}$ by 19.0\% and $S_{\text{sem}}$ by 7.6\%, showing that explicit 3D perception is essential for providing structured geometric inputs to DQNet and forming reliable object- and scene-level LiDAR representations.

Dual querying branches play complementary roles in DQNet. Removing object branch reduces $S_{\text{num}}$ by 4.0\%, indicating its contribution to instance-level grounding through object positions and local spatial relations. Removing scene branch causes a larger drop of 9.5\%, indicating that global BEV context is crucial for lane topology, intersection layout, and long-range spatial reasoning. This explains why the dual-branch design mainly improves numerical QA, where accurate 3D spatial evidence is required.

LDA is the key interface between LiDAR representations and language reasoning. Removing LDA blocks decoder-level access to DQNet output and degrades the model to a vision-language baseline incapable of effectively capturing LiDAR information, leading to a 26.7\% drop in $S_{\text{num}}$ and a 11.1\% drop in $S_{\text{sem}}$. This indicates that encoding and aligning LiDAR features alone is insufficient. They must be explicitly injected into decoder layers to support final language reasoning.
\begin{table}[t]
\small
\centering
\begin{tabular}{c c c c}
\toprule
Backbone
& $S_{\text{text}} \uparrow$ 
& $S_{\text{num}} \uparrow$
& $S_{\text{sem}} \uparrow$ \\
\midrule
Qwen3-VL-2B-Instruct &  0.853 & 0.558 & 0.812 \\
Qwen3-VL-4B-Instruct &  0.859 & 0.589 & 0.828 \\
Qwen3-VL-8B-Instruct &  0.869 & 0.617 & 0.848 \\
\bottomrule
\end{tabular}
\caption{Effect of backbone model scale on Inter-Geo.}
\label{tab:backbone}
\end{table}

\textbf{Effect of Different Backbone Scales.} We evaluate Inter-Geo with backbones of different scales on grounded Inter-3D VQA. As shown in Table~\ref{tab:backbone}, scaling the backbone from 4B to 8B improves $S_{\text{text}}$, $S_{\text{num}}$, and $S_{\text{sem}}$ by 1.2\%, 4.7\%, and 2.4\%, respectively. Notably, even the 2B variant outperforms existing VLM baselines, suggesting that aligned LiDAR representations provide complementary 3D geometric cues beyond standard VLM capabilities. These results show that Inter-Geo benefits from stronger backbones while retaining our advantage in multimodal LiDAR integration design.

\begin{table}[t]
\small
\centering
\begin{tabular}{l l c c c}
\toprule
Injection & Stage
& $S_{\text{text}} \uparrow$ 
& $S_{\text{num}} \uparrow$
& $S_{\text{sem}} \uparrow$ \\
\midrule
Input Embed & Input & 0.856 & 0.552 & 0.816\\
Vision Fusion & Input & 0.850 & 0.549 & 0.810 \\
LDA & Decoder & 0.859  & 0.589 & 0.828 \\
\bottomrule
\end{tabular}
\caption{Effect of LiDAR modality injection strategy.}
\label{tab:lidar_injection}
\end{table}

\begin{table*}[t]
\centering
\begin{tabular}{lccc}
\hline
\textbf{Model} & \textbf{$S_{\text{text}} \uparrow$} & \textbf{$S_{\text{num}} \uparrow$} & \textbf{$S_{\text{sem}} \uparrow$} \\
\hline
Qwen3-4B & 0.769 (-10.5\%) & 0.365 (-38.0\%) & 0.510 (-38.4\%) \\
Qwen3-8B & 0.783 (-8.8\%)  & 0.411 (-30.2\%) & 0.573 (-30.8\%) \\
\hline
\end{tabular}
\caption{Performance of text-only LLMs on the grounded Inter-3D VQA, with relative performance drops compared to Inter-Geo.}
\label{tab:text_only_llm}
\end{table*}

\textbf{Effect of LiDAR Modality Injection Strategy.} In Table~\ref{tab:lidar_injection}, we compare two alternative LiDAR injection strategies with LDA. The input-embed strategy appends LiDAR embeddings to the input sequence as \texttt{<image\_embeds>} + \texttt{<lidar\_embeds>} + \texttt{<text\_tokens>}. Although this improves over image-only VLMs, it treats LiDAR features as ordinary tokens, forcing geometric cues to compete with image and text tokens in the same attention space. The vision fusion strategy injects LiDAR into image embeddings through cross-attention, forming \texttt{<lidar-enhanced image\_embeds>} + \texttt{<text\_tokens>}. This early-fusion strategy may disrupt pretrained visual representations and dilute independent object- and scene-level geometric structures. In contrast, LDA allows the decoder to selectively retrieve 3D cues conditioned on language states, preserving pretrained image--text alignment while improving semantic grounding and spatial reasoning.

\begin{table}[t]
\centering
\begin{tabular}{lc}
\hline
\textbf{Model} & \textbf{Inference Speed} \\
\hline
Qwen3-4B       & 3.1 FPS \\
Qwen3-VL-4B    & 2.6 FPS \\
Inter-Geo      & 1.3 FPS \\
\hline
\end{tabular}
\caption{Inference speed comparison of Inter-Geo with text-only LLM and image-only VLM baselines.}
\label{tab:runtime}
\end{table}

\textbf{Effect of Visual Information.} We additionally evaluate text-only LLMs on the grounded Inter-3D VQA to assess the contribution of visual information. As shown in Table~\ref{tab:text_only_llm}, text-only models consistently underperform vision-language models and Inter-Geo, with particularly large performance gaps on numerical and semantic metrics. These results indicate that language priors alone are insufficient for grounded roadside reasoning and highlight the importance of incorporating visual and geometric information.

\textbf{Runtime and Deployment Analysis.} To assess inference speed, we compare Inter-Geo with text-only and image-only models using the same backbone as shown in Table~\ref{tab:runtime}. All models are evaluated on a single GPU with batch size 1. While Inter-Geo incurs additional computational cost, the text-only baseline reaches only 3.1 FPS, indicating that inference efficiency remains a broader challenge. In practice, sensing, preprocessing, and data transmission further contribute to end-to-end latency, motivating future work on more efficient deployment.

\section{Conclusion}
We introduce Inter-3D VQA, a roadside multimodal benchmark integrating multi-view images, global point clouds, and infrastructure topology for spatiotemporal reasoning. We further develop Inter-Geo, a LiDAR-enhanced MLLM that serves as a baseline for multimodal roadside reasoning. Experiments demonstrate the value of incorporating explicit LiDAR geometry, particularly for grounded spatial and numerical reasoning, while controlled ablations evaluate the contribution of the proposed fusion design. We hope Inter-3D VQA provides a foundation for future research on 3D-aware MLLMs for roadside traffic understanding.

\section*{Limitations}
Although we introduce Inter-3D VQA, a large-scale roadside multimodal benchmark for 3D reasoning, and Inter-Geo, a multimodal large language model that incorporates LiDAR information with multi-view images, there are still several limitations.

First, while the results suggest that the 3D VQA tasks introduced in our benchmark are challenging for current models, the data underlying the benchmark is collected from a single intersection with a fixed geometry and layout. Cross-intersection validation is therefore necessary in future work. Second, Inter-Metrics depends on structured parsing of model outputs. When responses deviate from expected formats, semantic parsing may fail and affect $S_{\text{sem}}$, even if part of the answer is reasonable. Third, Inter-Geo mainly uses single-frame multimodal inputs and does not fully exploit multi-frame temporal cues, limiting its ability to model long-term motion patterns and dynamic interactions. Future work will expand scenario diversity, improve robust semantic parsing, and develop multi-frame LiDAR-image reasoning models.

\section*{Ethical Considerations}
As MLLMs are increasingly explored for ITS applications, greater attention should be given to how their outputs are interpreted and used in practice. Inaccurate outputs may lead to suboptimal policy or operational decisions, such as unnecessary signal-timing changes, or overly conservative roadside safety warnings. Model performance may also vary across regions with different traffic rules, roadway designs, and population characteristics. These considerations highlight the importance of context-specific validation and appropriate use of MLLMs as decision-support tools.

\bibliography{custom}

\clearpage

\appendix
\section*{Appendix}
\section{Inter-3D VQA Dataset}

\subsection{Task Definition}
While the main paper introduces the overall task formulation, we provide additional details and representative examples here to better illustrate the scope and challenges of each task category.

Existing roadside VQA benchmarks~\cite{TUMTrafficVideoQA,RoadSceneVQA} are limited to image-based semantic understanding, which restricts their ability to support fine-grained spatial and structural reasoning. For example, questions such as “How far is a pedestrian from the nearest crosswalk?” or “What is the distance between a vehicle and the stop line?” require precise geometric measurements that are difficult to estimate reliably from 2D images alone. Similarly, queries like “How many queued vehicles are in the northbound left-turn lane?” or “Is there a pedestrian at the southeast curb ramp?” involve topology-aware reasoning, requiring both spatial localization and an understanding of infrastructure semantics. In addition, interaction-based questions, such as identifying potential conflicts or unsafe behaviors among traffic participants, require modeling dynamic relationships over time. To address these challenges, Inter-3D VQA extends image-based reasoning to multimodal reasoning over synchronized multi-view images and global point clouds. This enables models to perform unified reasoning over 3D geometry, infrastructure topology, and temporal dynamics within a consistent spatial coordinate system.

We categorize the benchmark into four complementary task types:
\begin{itemize}
    \item Basic Perception: This category focuses on a model's ability to identify fundamental information, including object types, counts, visual attributes, and environmental conditions. While image-based models typically perform well on such tasks, incorporating point cloud data provides more spatial observations especially under occlusion or limited viewpoints.
    \item Spatial Reasoning: This category emphasizes the evaluation of a model's geometric understanding within a three-dimensional space. The model is required to estimate absolute distances between objects using both images and point clouds, while also understanding the topological relationships between objects and infrastructure elements (e.g., lanes and crosswalks) for more precise spatial reasoning.
    \item Temporal Reasoning: Using sequential multi-frame images and point clouds, this category evaluates a model's ability to capture dynamic traffic scenarios, including motion analysis and trajectory inference for individual objects, as well as interaction modeling among multiple objects (e.g., following distance estimation and conflict detection).
    \item Scene Understanding: From a global perspective, this category integrates multi-view images and global point clouds to evaluate a model’s understanding of overall intersection traffic states, including congestion analysis, risk detection, and traffic management.
\end{itemize}

\begin{figure*}[ht]
    \centering 
    \includegraphics[width=\linewidth]{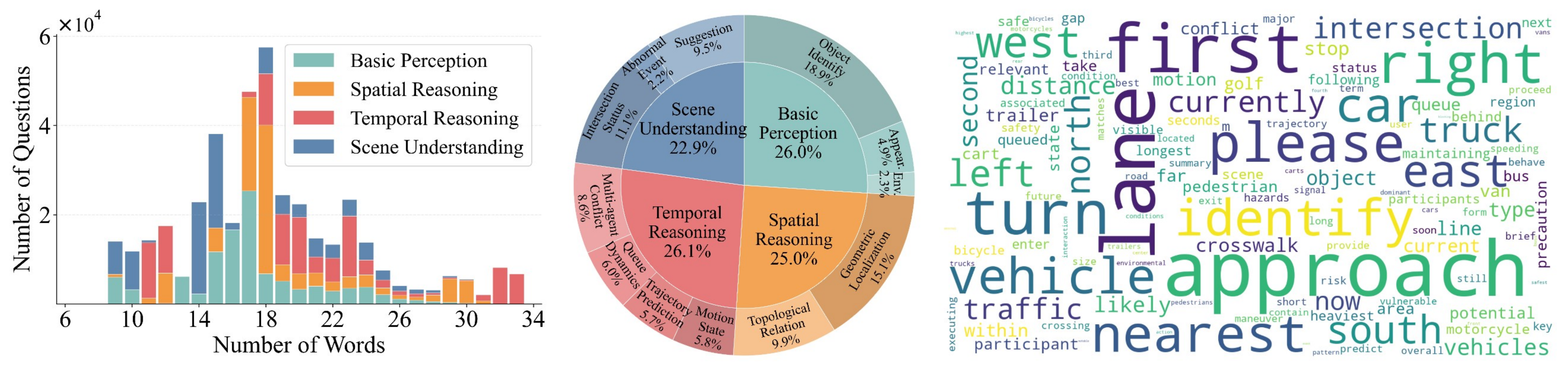}
    \caption{Question-level statistics of Inter-3D VQA. The dataset exhibits a balanced distribution across task categories and provides rich topology-aware descriptions for 3D reasoning.}
    \label{fig:qa_statistics}
\end{figure*}

\begin{figure*}[ht]
\centering
\begin{subfigure}{0.49\linewidth}
    \centering
    \includegraphics[width=\linewidth]{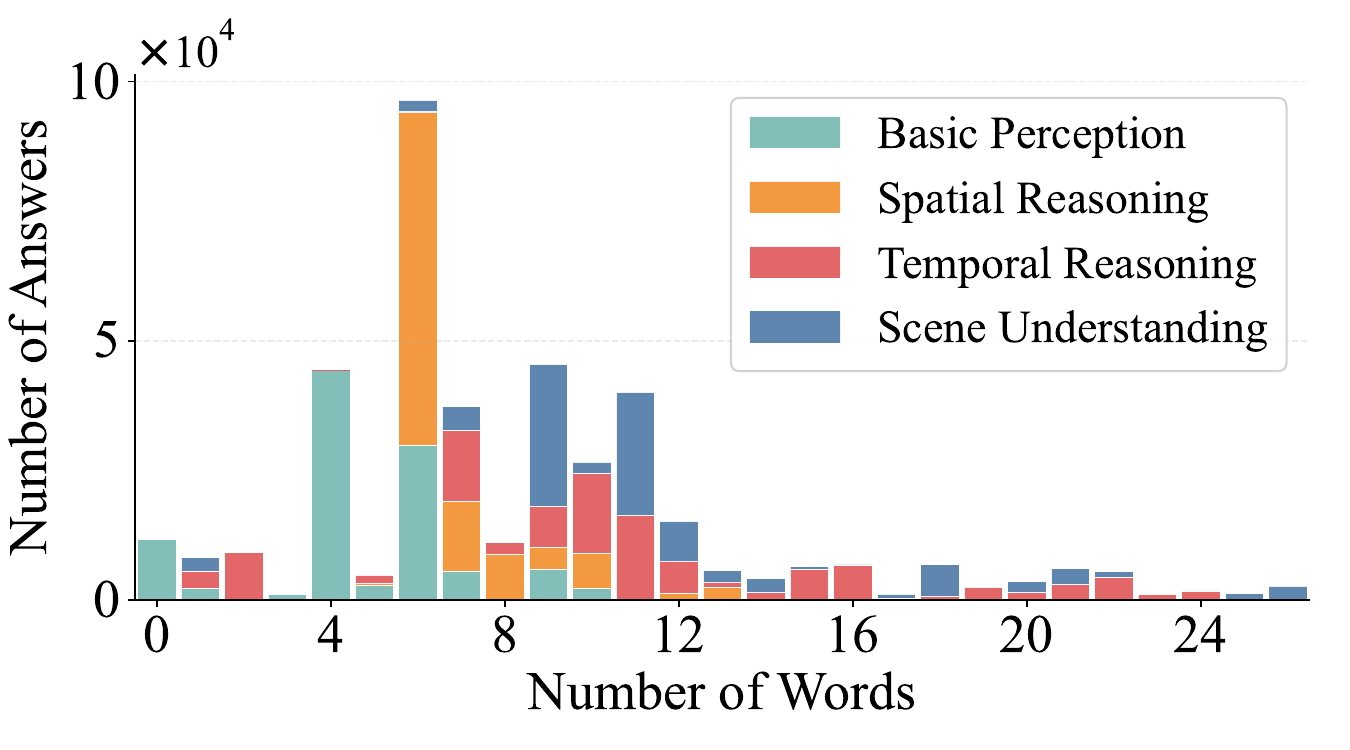}
    \caption{}
        \label{fig:answer_length}
\end{subfigure}
\hfill
\begin{subfigure}{0.49\linewidth}
    \centering
    \includegraphics[width=\linewidth]{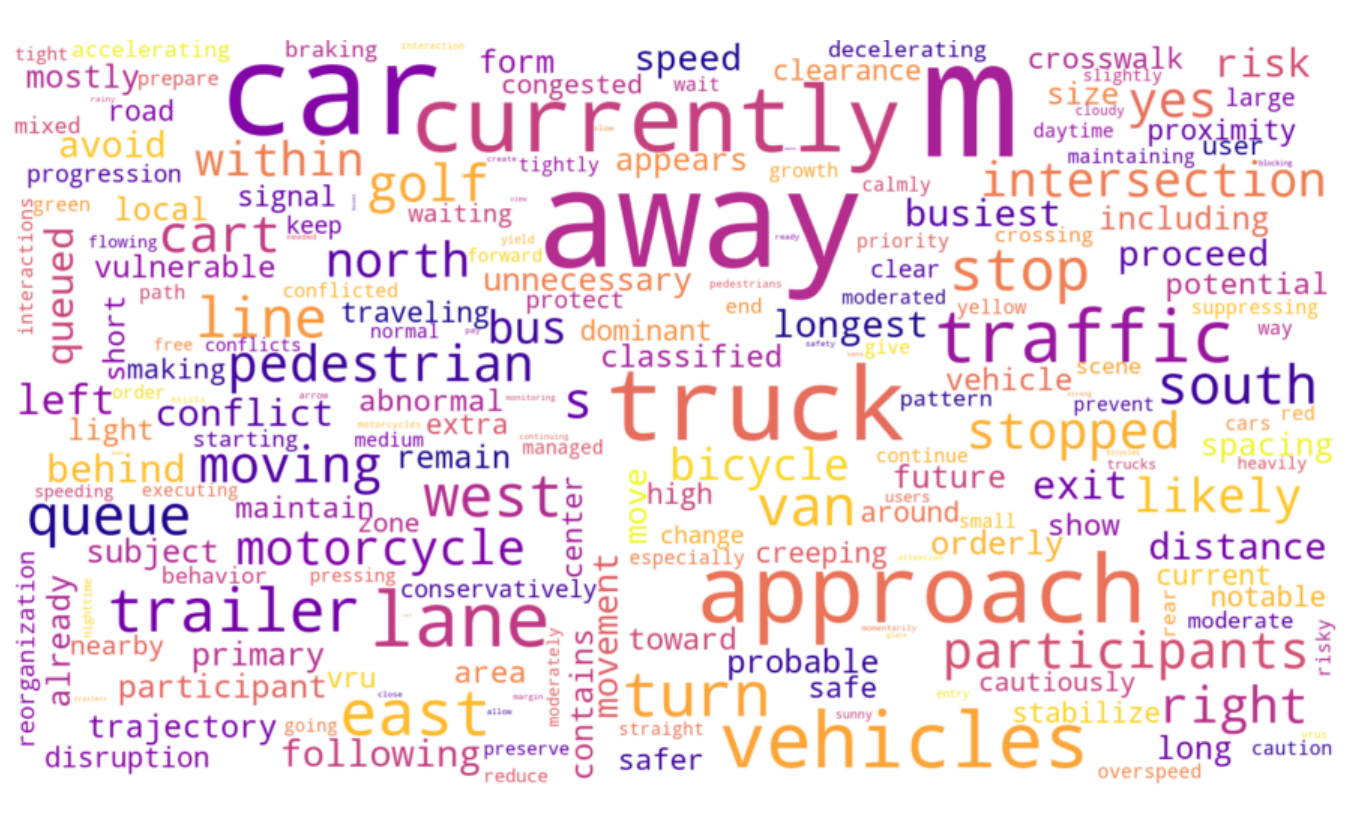}
    \caption{}
    \label{fig:answer_wordcloud}
\end{subfigure}
\caption{Answer-level statistics of Inter-3D VQA. (a) Distribution of answer lengths, demonstrating concise and reasoning-oriented responses. (b) Word cloud of answers, revealing dominant semantic concepts in traffic participants, infrastructure, and motion.}
\label{fig:answer}
\end{figure*}

\subsection{Dataset Statistics}
As shown in Fig.~\ref{fig:qa_statistics}, we analyze Inter-3D VQA from multiple perspectives, including question length distribution, task composition, and vocabulary statistics. The histogram of question lengths shows that most questions contain approximately 14--22 words, indicating a balance between concise formulation and sufficient semantic detail for complex reasoning. The task distribution demonstrates a relatively balanced composition across the four task categories, reducing bias toward specific reasoning patterns. In particular, Spatial Reasoning and Temporal Reasoning occupy a substantial proportion of the benchmark, highlighting the emphasis on geometry-aware and dynamic traffic understanding. 

The vocabulary statistics further reveal rich semantic coverage related to traffic participants, infrastructure topology, motion states, and interaction patterns. High-frequency terms such as ``lane'', ``crosswalk'', ``turn'', and directional descriptors indicate that the dataset explicitly models topology-aware and infrastructure-centric reasoning.

We further examine the distribution of answer lengths and their semantic content, assess object coverage, and analyze the distribution of numerical attributes. As shown in Fig.~\ref{fig:answer_length}, the distribution of answer lengths reveals that most responses remain concise, indicating that the benchmark prioritizes precise argumentation and reasoning rather than the generation of verbose answers. The word cloud in Fig.~\ref{fig:answer_wordcloud} highlights high-frequency terms related to traffic participants, road structures, motion states, and decision-making cues, underscoring the close interplay between perception and reasoning within intersection scenarios.

\begin{figure*}[ht]
\centering
\begin{subfigure}{0.49\linewidth}
    \centering
    \includegraphics[width=\linewidth,height=0.5\linewidth]{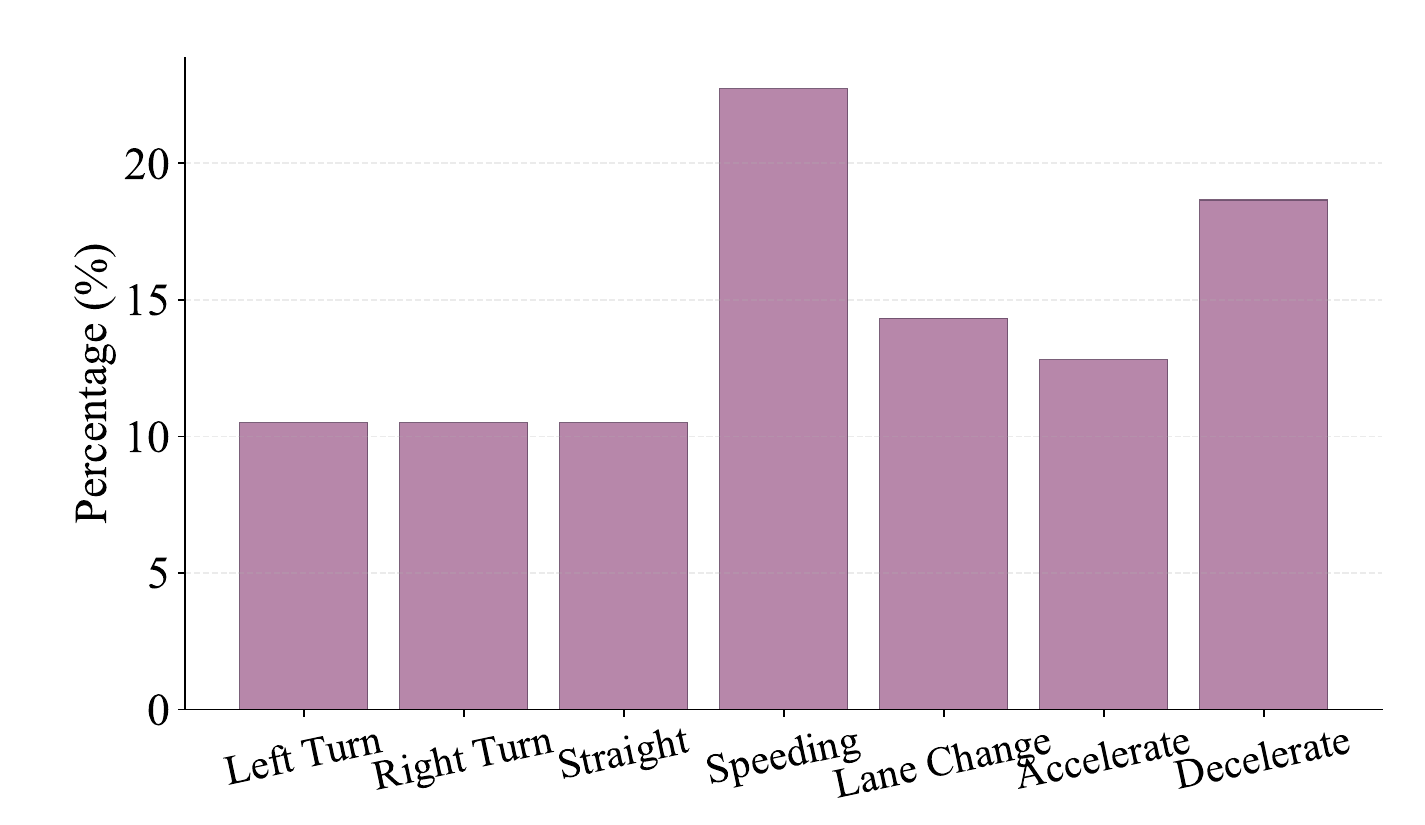}
    \caption{}
    \label{fig:behavior_proportions}
\end{subfigure}
\hfill
\begin{subfigure}{0.49\linewidth}
    \centering
    \includegraphics[width=\linewidth,height=0.5\linewidth]{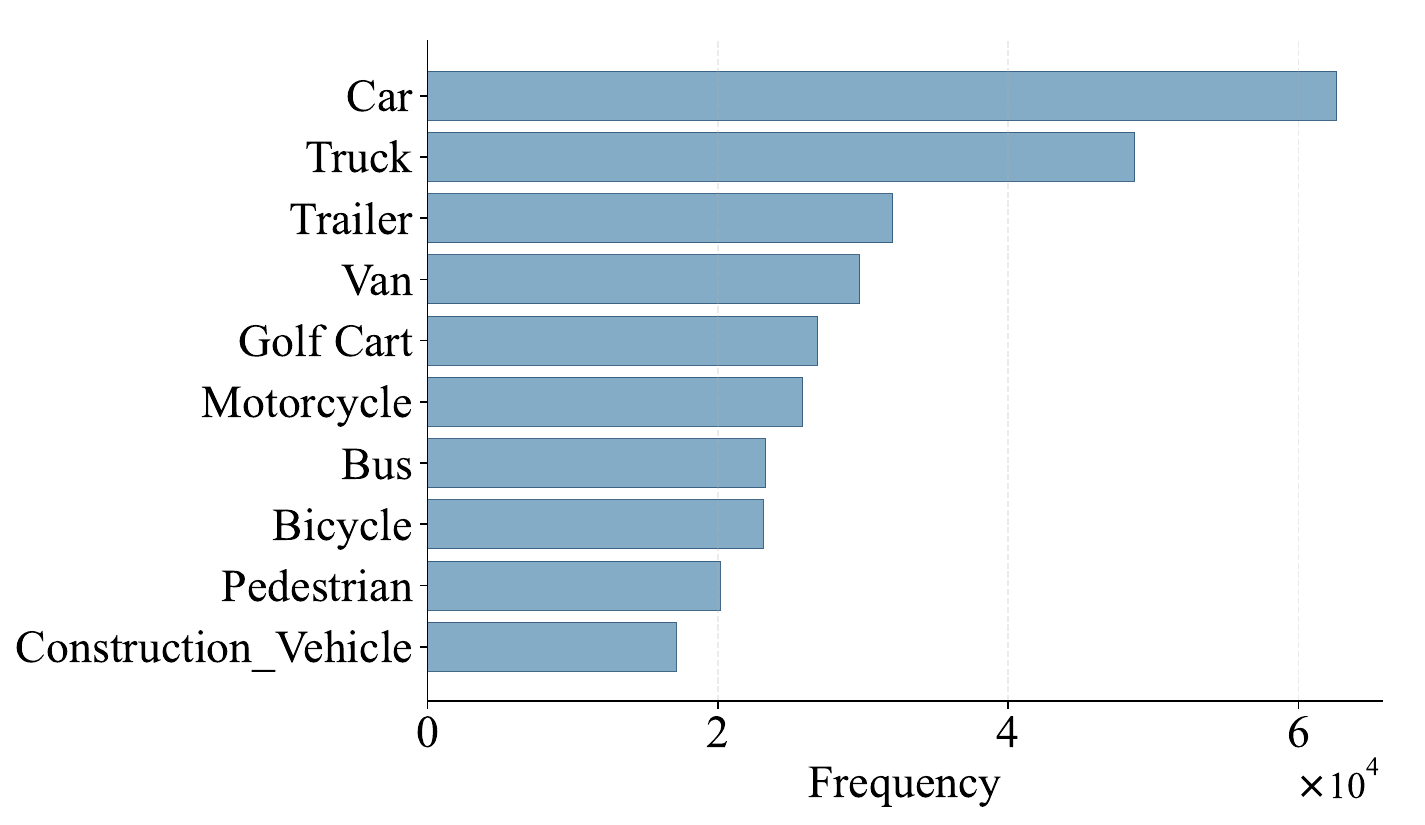}
    \caption{}
    \label{fig:object_category}
\end{subfigure}
\caption{Behavior and object statistics of Inter-3D VQA. (a) Distribution of driving behaviors, covering both routine and safety-critical actions. (b) Distribution of object categories, including diverse traffic participants.}
\label{fig:behavior_object}
\end{figure*}

\begin{figure*}[ht]
\centering
\includegraphics[width=\linewidth]{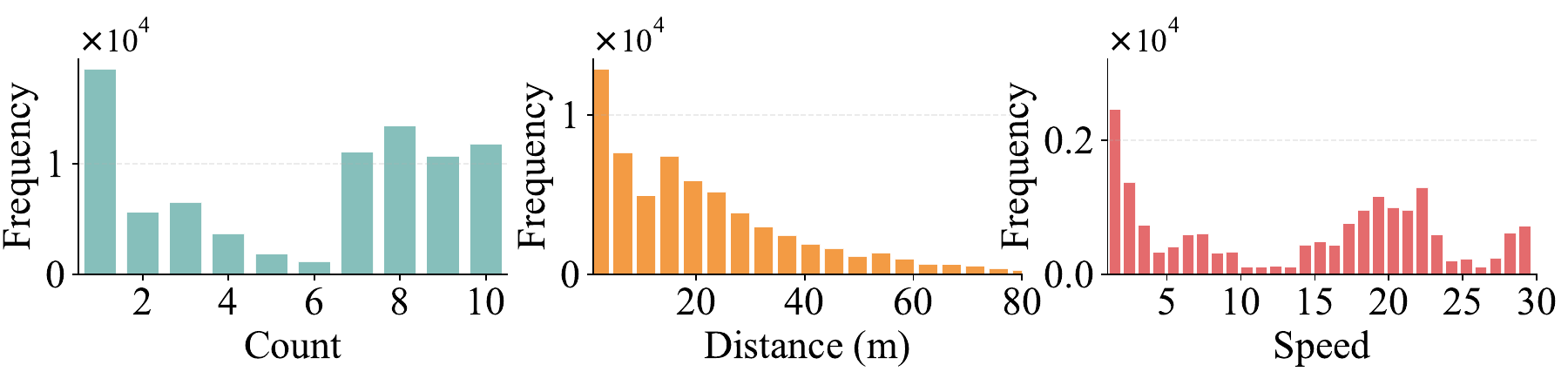}
\caption{Distribution of numerical attributes in Inter-3D VQA, including object count, distance, and speed.}
\label{fig:numeric_value}
\end{figure*}

\begin{figure}[ht]
\centering
\includegraphics[width=\linewidth]{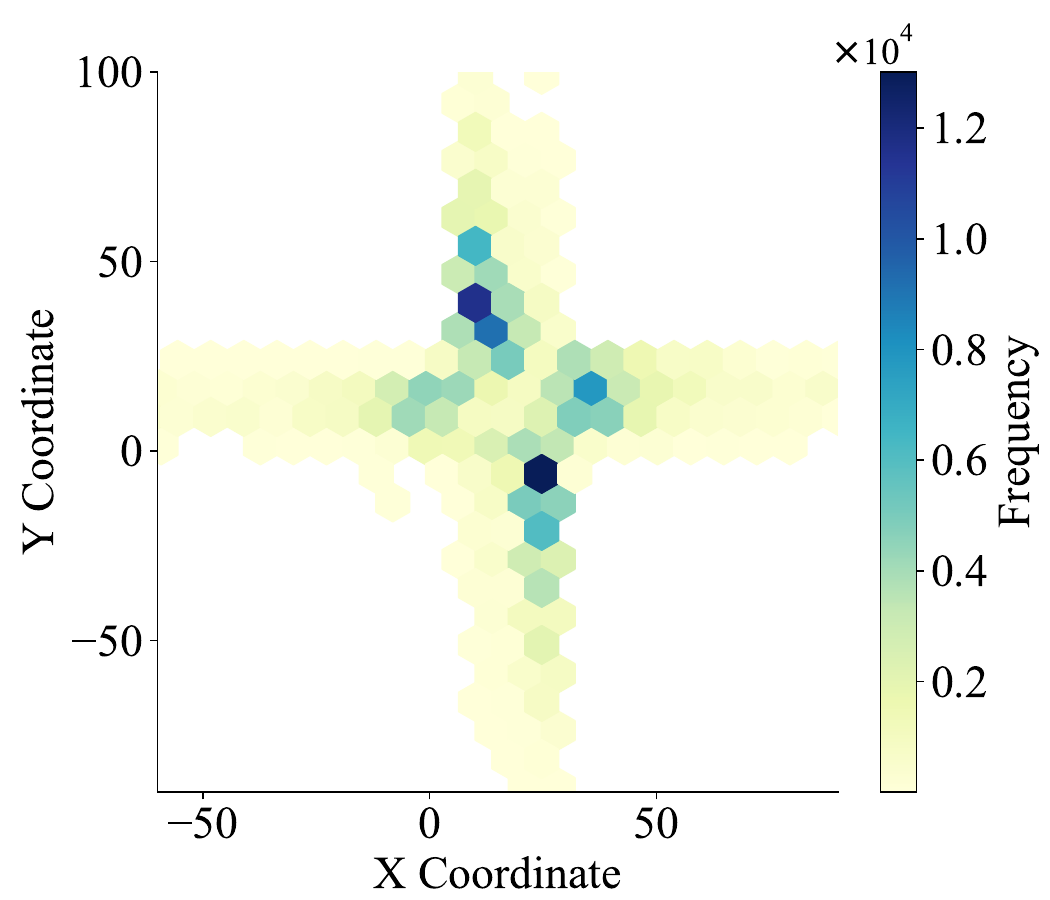}
\caption{Spatial distribution of 3D coordinates in Inter-3D VQA.}
\label{fig:coordinate}
\end{figure}

Fig.~\ref{fig:behavior_proportions} shows a diverse distribution of driving behaviors, including turning, lane changes, acceleration, and speeding, enabling evaluation across both routine and safety-critical situations. Meanwhile, Fig.~\ref{fig:object_category} demonstrates broad object coverage, including common vehicles as well as vulnerable and less frequent traffic participants, supporting diverse and realistic evaluation settings.

We also analyze the distributions of numerical targets and 3D spatial coordinates in QA pairs. As shown in Fig.~\ref{fig:numeric_value}, statistics of counts, distances, and velocities indicate that the dataset provides rich quantitative supervision, supporting tasks such as quantity estimation, geometric reasoning, and motion analysis. Fig.~\ref{fig:coordinate} shows that the distribution of 3D coordinates involved in QA pairs spans a wide range of intersection areas, indicating a wide spatial coverage rather than concentration in limited regions.

\subsection{Inter-Metrics Details}
We provide additional implementation details and evaluation procedures for Inter-Metrics. The framework is designed to jointly evaluate linguistic quality, geometric reasoning, and semantic correctness in Inter-3D VQA tasks.

\textbf{Textual Consistency.} We evaluate the generated answers using a weighted combination of lexical and semantic similarity metrics: Normalized Exact Match (NEM), BLEU-4~\cite{BLEU}, ROUGE-L~\cite{rouge}, BERTScore~\cite{BERTScore}, and SimCSE~\cite{SimCSE} similarity. For reproducibility, we specify the implementation details and parameter settings for each metric. BLEU-4 is computed using \texttt{sacrebleu==2.4.0} with the 13a tokenizer, case-sensitive scoring, exponential smoothing, and corpus-level aggregation. ROUGE-L is computed using \texttt{rouge-score==0.1.2}, and we report ROUGE-L F1 with \texttt{use\_stemmer=True}. BERTScore is computed using \texttt{bert-score==0.3.13}, and we report the F1 score with roberta-large, \texttt{num\_layers=17}, \texttt{idf=False}, and \texttt{rescale\_with\_baseline=True}. SimCSE similarity is computed using the \texttt{princeton-nlp/sup-simcse-roberta-large} checkpoint, where predictions and references are embedded separately and compared using cosine similarity. The final score is averaged over all evaluation examples.
 
Compared with conventional language-based evaluation, Inter-Metrics places greater emphasis on semantic similarity to reduce sensitivity to wording variations. For example, “The vehicle is turning left” and “The car is making a left turn” may receive relatively low scores under lexical-overlap metrics such as BLEU or ROUGE, despite describing the same traffic behavior. By incorporating semantic similarity metrics such as BERTScore and SimCSE, Inter-Metrics better captures semantic equivalence.

\textbf{Numerical Accuracy.} Unlike image-based spatial evaluation used in previous roadside benchmarks, Inter-Metrics evaluates quantities directly in physically meaningful 3D space. Scalar quantities (e.g., object count, distance, speed, and acceleration) are evaluated using absolute error, while vector quantities (e.g., 3D coordinates, image-plane coordinates, and trajectory waypoints) are evaluated using Euclidean distance error. To ensure comparability across heterogeneous quantities, each error is normalized using a predefined threshold specific to the corresponding field. Thresholds are set to 2 for count, 10 m for distance, 5 m/s for speed, 1 m/s$^2$ for acceleration, 8 m for global 3D coordinates, 150 px for image-plane coordinates, and 5 m for waypoint offsets. This allows the framework to more reliably measure geometric understanding, spatial localization, and trajectory reasoning. Although metric calculation relies on structured output parsing, all models are supervised to follow a unified answer format, resulting in a high parsing success rate.

\textbf{Semantic Correctness.} Inter-Metrics additionally evaluates structured semantic fields extracted from model predictions, including object categories, lane functions, signal states, motion directions, interaction patterns, and risk-related attributes. Each field is evaluated through exact matching against the ground-truth structured annotations. This design explicitly distinguishes semantic correctness from surface-level linguistic similarity, preventing semantically incorrect yet linguistically plausible predictions from receiving overly high scores. For example, “A vehicle on the northbound approach is turning left” and “A vehicle on the southbound approach is turning left” may obtain high lexical-overlap scores despite describing different traffic situations. Inter-Metrics explicitly evaluates structured semantic fields such as direction and interaction state, preventing semantically inconsistent predictions from being overestimated.

\section{Implementation Details}
All models are evaluated on the same Inter-Metrics protocol. We use the same train/validation split across methods, with scene-level splitting and identical evaluation scripts. All reported results are averaged over five independent runs. Input camera order is fixed to \texttt{north}, \texttt{south}, \texttt{east}, and \texttt{west}.
\subsection{Inter-Geo}
We provide the main architectural settings of Inter-Geo. The model uses two LiDAR input branches: object-level features and scene-level LiDAR features. Raw object and scene features have dimensions 523 and 386, respectively, and are projected into a shared LiDAR hidden space of dimension 512. For numeric object attributes, we additionally employ label embeddings with vocabulary size 16 and embedding dimension 32.

To regulate LiDAR information injection, Inter-Geo retains at most 16 object representations, while scene-level inputs are truncated to 256 tokens and compressed into 64 scene representations before interacting with the decoder. Cross-modal attention in the LiDAR branch uses 8 attention heads. The top 4 decoder layers are further equipped with LiDAR Decoder Adapters (LDA), each using 8 attention heads. To stabilize early training, the adapter gates are initialized with bias $-5.0$, suppressing excessive LiDAR influence at initialization.

For optimization, Inter-Geo adopts LoRA fine-tuning on the Qwen3-VL-4B backbone with rank 8, scaling factor 16, and dropout 0.05. The image resolution is capped at 262,144 pixels, and the maximum sequence length is set to 4096.

\subsection{Open-source Generalist VLMs}
All open-source VLMs are trained with BF16 precision using a cosine learning rate schedule and a base learning rate of $1\times10^{-4}$, taking four-view images as input. Qwen3-VL~\cite{Qwen3-VL} and InternVL~\cite{InternVL3} models use a maximum image resolution of 262,144 pixels and sequence length of 2,048. The 4B variants adopt standard LoRA with rank 8, while the 8B variants employ rsLoRA with rank 32, $\alpha=64$, and dropout 0.05. These models are trained with batch size 2 and gradient accumulation 4. 

LLaVA-NeXT-8B~\cite{LLaVA-NeXT} uses a lower image resolution cap of 196,608 pixels and a longer sequence length of 8,192, while Llama-3.2-11B-Vision-Instruct~\cite{llama32} uses a resolution of 262,144 pixels and sequence length of 2,048. Both models are trained with batch size 1 and gradient accumulation 8.

\subsection{Specialized Driving VLMs}
We follow the original training and inference protocols of all specialized driving VLMs, adapting only the data interfaces for Inter-3D VQA. DriveLM~\cite{drivelm} retains its LLaMA-Adapter-based 7B multimodal backbone and is trained for 3 epochs with batch size 4, effective learning rate $6.25\times10^{-5}$, and weight decay 0.05. Senna~\cite{Senna} adopts LoRA fine-tuning for 3 epochs using batch size 1, gradient accumulation 8, and learning rate $1\times10^{-4}$. NuScenes-QA~\cite{NuScenesQA} preserves its MCAN-style~\cite{MCAN} structured reasoning architecture and object-centric pipeline, trained for 4 epochs with batch size 32 and learning rate $3\times10^{-4}$.

OmniDrive~\cite{OmniDrive} maintains its camera-only multi-view setting with four fixed views and is trained for 3 epochs using global batch size 12 and learning rate $2\times10^{-5}$. OpenDriveVLA~\cite{OpenDriveVLA} preserves its original feature extraction and language generation pipeline with fixed camera ordering, trained using batch size 1, gradient accumulation 8, and learning rate $1\times10^{-4}$. BEV-LLM retains its ``BEV Features $\rightarrow$ Q-Former $\rightarrow$ LLM'' architecture and applies LoRA fine-tuning with rank 16, batch size 2, and learning rate $1\times10^{-4}$ for 3 epochs.

\begin{table*}[t]
\centering
\begin{tabular}{lcccccc}
\toprule
\multirow{2}{*}{\textbf{Models}} & \multicolumn{3}{c}{\textbf{Free-form QA}} & \multicolumn{3}{c}{\textbf{Grounded QA}} \\
\cmidrule(lr){2-4} \cmidrule(lr){5-7}
& \textbf{$S_{\text{text}} \uparrow$} & \textbf{$S_{\text{num}} \uparrow$} & \textbf{$S_{\text{sem}} \uparrow$}
& \textbf{$S_{\text{text}} \uparrow$} & \textbf{$S_{\text{num}} \uparrow$} & \textbf{$S_{\text{sem}} \uparrow$} \\
\midrule

GPT-5.1~\citeyearpar{gpt51}            & 0.715 & \underline{0.251} & 0.407 & 0.523 & 0.081 & 0.098 \\
Gemini-2.5-flash~\citeyearpar{gemini25flash} & \underline{0.725} & 0.204 & \underline{0.428} & 0.534 & \underline{0.141} & \underline{0.125} \\
Claude-4.5-Haiku~\citeyearpar{claudehaiku45}   & 0.669 & 0.197 & 0.335 & \underline{0.537} & 0.075 & 0.077 \\

\bottomrule
\end{tabular}
\caption{Zero-shot performance of closed-source API-only VLMs on free-form and grounded Inter-3D VQA.}
\label{tab:api_vlm_results}
\end{table*}

\begin{table*}[t]
\centering
\begin{tabular}{l l c c c}
\toprule
\textbf{Model} &
\textbf{Backbone}
& $S_{\text{text}} \uparrow$ 
& $S_{\text{num}} \uparrow$
& $S_{\text{sem}} \uparrow$ \\
\midrule
CenterPoint~\cite{centerpoint} & Sparse Convolution & 0.862 & 0.540 & 0.828 \\
TransFusion-L~\cite{TransFusion} &  Sparse Convolution & 0.857 & 0.580 & 0.819\\
DSVT~\cite{DSVT} & Transformer &  0.855 & 0.544 & 0.812 \\
LION~\cite{LION} & Mamba & 0.859 & 0.589 & 0.828 \\
\bottomrule
\end{tabular}
\caption{Comparison of different 3D encoders.}
\label{tab:3d_encoder}
\end{table*}

\subsection{Specialized Roadside VLMs}
TraffiX-Qwen~\cite{TUMTrafficVideoQA} adopts a 3-frame $\times$ 4-view configuration, yielding 12 images per QA sample. It uses Qwen2-0.5B with a SigLIP~\cite{SigLIP} vision encoder as the backbone and is trained for 3 epochs using LoRA-based adaptation, with batch size 1, gradient accumulation 4, and learning rate $5\times10^{-6}$.

\subsection{Closed-source API-only VLMs}
Closed-source models are evaluated in a pure zero-shot setting without fine-tuning, using the same input samples and evaluation scripts as open-source models. We evaluate GPT-5.1~\cite{gpt51}, Gemini-2.5-Flash~\cite{gemini25flash}, and Claude Haiku 4.5~\cite{claudehaiku45}, all of which take four input images per sample. GPT uses low-detail image mode with reasoning disabled and a maximum of 512 output tokens. Gemini adopts temperature 0, \texttt{MEDIA\_RESOLUTION\_LOW}, and 512 maximum output tokens, with \texttt{thinkingBudget} set to 0 for Gemini-2.5-Flash. Claude uses temperature 0, resizes images to a maximum long edge of 1280 pixels, and limits outputs to 128 tokens.

\section{Benchmark Results}

\subsection{Evaluation of API-only VLMs}
We report the zero-shot performance of closed-source API-only VLMs using four-view image inputs in Table~\ref{tab:api_vlm_results}. Since these models are evaluated without supervised fine-tuning, they are not adapted to the answer formats and structured fields required by our benchmark. As a result, although they show moderate language understanding on free-form QA, their performance drops substantially on grounded QA. For free-form QA, Gemini-2.5-Flash~\cite{gemini25flash} achieves the best textual and semantic scores, while GPT-5.1~\cite{gpt51} obtains the highest numerical score. However, on grounded QA, the best numerical and semantic scores remain only 0.141 and 0.125, respectively. 

This degradation is mainly caused by two factors. First, without task-specific training, API-only models struggle to follow the required structured response format, especially for grounded answers involving object references, coordinates, numerical values, and semantic fields. Second, their outputs are often more free-form and descriptive, which increases parsing failures in numerical and semantic evaluation. Therefore, even when a response contains partially relevant information, it may not be successfully parsed into the expected fields, leading to low $S_{\text{num}}$ and $S_{\text{sem}}$.

\subsection{Different 3D Encoders}
We compare different 3D encoders within the Inter-Geo framework in Table~\ref{tab:3d_encoder}. All variants outperform image-based VLM baselines, demonstrating that the proposed DQNet and LDA design is compatible with diverse LiDAR perception backbones while consistently preserving the advantage of explicit 3D modeling.

Among these encoders, the Mamba-based LION~\cite{LION} achieves the best overall performance, indicating its strength in modeling long-range spatial dependencies in point clouds. This capability is particularly beneficial for global scene understanding and spatial reasoning. TransFusion-L~\cite{TransFusion} obtains a strong numerical score ($S_{\text{num}}=0.580$), suggesting that its detection-oriented strategy and object-level representations are effective for distance estimation and localization, although this does not necessarily translate into the best structured semantic accuracy.

CenterPoint~\cite{centerpoint} achieves the highest textual score ($S_{\text{text}}=0.862$) but lower numerical performance ($S_{\text{num}}=0.540$), likely because sparse convolution is effective for object detection but less expressive for global geometric reasoning. DSVT~\cite{DSVT} remains competitive, yet its numerical and semantic scores are below LION, suggesting that Transformer-based voxel modeling is slightly less effective for global point cloud patterns in roadside scenarios.

\begin{table}[t]
\centering
\begin{tabular}{c c c c}
\toprule
\textbf{Insertion Position}
& $S_{\text{text}} \uparrow$ 
& $S_{\text{num}} \uparrow$
& $S_{\text{sem}} \uparrow$ \\
\midrule
First 4 layers & 0.858 & 0.554 & 0.816 \\
Middle 4 layers & 0.858 & 0.552 & 0.822\\
Last 4 layers & 0.859 & 0.589 & 0.828 \\
\bottomrule
\end{tabular}
\caption{Effect of LDA insertion position in the decoder on Inter-Geo.}
\label{tab:lda_position}
\end{table}

\subsection{LDA Insertion Position}
We study the effect of LDA insertion position in the 36-layer Qwen3-VL-4B-Instruct decoder in Table~\ref{tab:lda_position}. Compared with inserting LDA into the last four layers, inserting it into the first four layers reduces $S_{\text{num}}$ by 6.3\% and $S_{\text{sem}}$ by 1.4\%. This is likely because early decoder layers mainly capture lexical patterns and low-level multimodal alignment. Injecting LiDAR representations too early may disturb the pretrained image--text representations and force geometric information to pass through all subsequent transformations, thereby weakening structured 3D cues.

Inserting LDA into the middle layers slightly improves semantic correctness compared with early insertion, but still yields a 6.3\% lower numerical score than late insertion. At this stage, the decoder has begun to form higher-level representations, but its hidden states may not yet be sufficiently task-specific to retrieve the most relevant object- and scene-level LiDAR representations.

In contrast, the last four decoder layers are closer to answer generation and contain more task-conditioned semantic states. Injecting LiDAR representations at this stage allows the model to access geometric cues after the visual-language context has been formed, reducing interference with pretrained representations while directly supporting final reasoning.

\begin{table}[t]
\centering
\begin{tabular}{c c c c}
\toprule
\textbf{\# LDA Layers}
& $S_{\text{text}} \uparrow$ 
& $S_{\text{num}} \uparrow$
& $S_{\text{sem}} \uparrow$ \\
\midrule
2 & 0.858 & 0.544 & 0.821 \\
4 & 0.859 & 0.589 & 0.828 \\
8 & 0.857 & 0.553 & 0.816\\
\bottomrule
\end{tabular}
\caption{Effect of the number of LDA layers in the decoder on Inter-Geo.}
\label{tab:lda_layer_num}
\end{table}

\subsection{Number of LDA Layers} We further investigate the effect of applying LDA to different numbers of decoder layers. As shown in Table~\ref{tab:lda_layer_num}, the 4-layer configuration achieves the best overall performance. Compared with the 2-layer setting, it improves $S_{\text{num}}$ by 8.3\% and $S_{\text{sem}}$ by 0.9\%, suggesting that an insufficient number of layers limits the model's ability to acquire 3D cues, whereas an appropriate number of LDA layers provides sufficient capacity to effectively capture LiDAR representations and integrate geometric cues into the language reasoning process.

However, increasing the number of LDA layers to 8 does not bring further gains. Instead, $S_{\text{num}}$ decreases by 6.1\% and $S_{\text{sem}}$ by 1.4\%. This indicates that excessive LiDAR injection may disturb pretrained language--vision alignment and introduce redundant geometric cues across decoder layers, causing the model to over-rely on LiDAR features.
\begin{table}[t]
\centering
\begin{tabular}{c c c c}
\toprule
\textbf{\# Object Queries}
& $S_{\text{text}} \uparrow$ 
& $S_{\text{num}} \uparrow$
& $S_{\text{sem}} \uparrow$ \\
\midrule
8  & 0.861 & 0.569 & 0.827 \\
16 & 0.859 & 0.589 & 0.828 \\
24 & 0.858 & 0.545 & 0.818\\
\bottomrule
\end{tabular}
\caption{Effect of the number of object queries in DQNet.}
\label{tab:obj_query}
\end{table}

\begin{table*}[t]
\centering
\small
\renewcommand{\arraystretch}{1.1}
\begin{tabular}{lccccccc}
\toprule
\textbf{Models} 
& \textbf{Count} $\downarrow$
& \textbf{Distance} $\downarrow$
& \textbf{Speed} $\downarrow$
& \textbf{Acceleration} $\downarrow$
& \textbf{$\text{Coord}_{\text{3d}}$} $\downarrow$
& \textbf{$\text{Coord}_{\text{2d}}$} $\downarrow$
& \textbf{Waypoint} $\downarrow$\\
\midrule

GPT-5.1             & 0.95 & 0.91 & 1.00 & 1.00 & 0.78 & 0.93 & 0.86 \\
Gemini-2.5-flash    & 0.97 & 0.91 & 0.51 & 0.99 & 0.85 & 0.94 & 0.84 \\
Claude-4.5-Haiku    & 0.95 & 0.94 & 0.71 & 1.00 & 0.99 & 1.00 & 0.90 \\

\midrule
LLaVA-NeXT-8B       & 0.35 & 0.24 & 0.69 & 0.94 & 0.81 & 0.87 & 0.71 \\
Llama3.2-11B        & 0.37 & 0.21 & 0.63 & 0.84 & 0.81 & 0.86 & 0.74 \\
Qwen3-VL-4B         & 0.26 & 0.18 & 0.53 & 0.64 & 0.77 & 0.81 & 0.78 \\
Qwen3-VL-8B         & 0.22 & 0.15 & 0.37 & 0.73 & 0.73 & \underline{\textbf{0.73}} & 0.66 \\
InternVL3-4B        & 0.28 & 0.20 & 0.54 & 0.74 & 0.80 & 0.89 & 0.72 \\
InternVL3-8B        & 0.25 & 0.18 & 0.38 & \underline{\textbf{0.54}} & 0.77 & 0.78 & 0.71 \\

\midrule
Senna-VLM           & 0.28 & 0.24 & 0.61 & 0.63 & 0.81 & 0.87 & 0.74 \\
NuScenes-QA         & 0.89 & 0.83 & 0.67 & 1.00 & 1.00 & 0.97 & 0.76 \\
OmniDrive           & 0.32 & 0.69 & 0.50 & 0.91 & 0.94 & 0.90 & 0.75 \\
OpenDriveVLA        & 0.33 & 0.67 & 0.44 & 0.74 & 0.75 & 0.90 & 0.82 \\
DriveLM             & 0.34 & 0.72 & 0.45 & 0.97 & 0.95 & 0.94 & 0.73 \\
BEV-LLM             & 0.34 & 0.63 & 0.31 & 0.93 & 0.97 & 0.99 & 0.72 \\

\midrule
TraffiX-Qwen        & 0.37 & 0.69 & 0.49 & 1.00 & 0.91 & 0.89 & 0.80 \\
Inter-Geo (ours)    & \underline{\textbf{0.22}} & \underline{\textbf{0.15}} & \underline{\textbf{0.29}} & 0.59 & \underline{\textbf{0.48}} & 0.76 & \underline{\textbf{0.31}} \\

\bottomrule
\end{tabular}
\caption{Detailed numerical error comparison on grounded Inter-3D VQA across different numerical categories.}
\label{tab:model_category_results}
\end{table*}
\subsection{Number of Object Queries}
In Table~\ref{tab:obj_query}, we evaluate the effect of varying the number of object queries in the object branch of DQNet. Reducing the number of queries to 8 mainly affects numerical reasoning, with $S_{\text{num}}$ decreasing by 3.4\%, suggesting that too few queries limit the model’s capacity to capture instance-level geometric cues.

Increasing the number of queries to 24 further degrades performance, reducing $S_{\text{num}}$ by 7.5\% and $S_{\text{sem}}$ by 1.2\%. In complex roadside scenes, excessive queries may attend to duplicated or low-relevance objects, introducing redundant or noisy representations and weakening the compactness of geometric evidence. Therefore, 16 object queries provide the best balance between object coverage and representation quality.

\begin{table}[t]
\centering
\begin{tabular}{c c c c}
\toprule
\textbf{\# Scene Queries}
& $S_{\text{text}} \uparrow$ 
& $S_{\text{num}} \uparrow$
& $S_{\text{sem}} \uparrow$ \\
\midrule
32  & 0.859 & 0.567 & 0.824 \\
64  & 0.859 & 0.589 & 0.828 \\
128 & 0.858 & 0.545 & 0.820\\
\bottomrule
\end{tabular}
\caption{Effect of the number of scene queries in DQNet.}
\label{tab:scene_query}
\end{table}

\subsection{Number of Scene Queries}
We further analyze the effect of the number of scene queries in the scene branch of DQNet. As shown in Table~\ref{tab:scene_query}, the default setting with 64 queries achieves the best balance between global context coverage and representation compactness.

Reducing the number to 32 decreases $S_{\text{num}}$ by 3.7\% and $S_{\text{sem}}$ by 0.5\%, suggesting that too few scene queries provide insufficient capacity to aggregate global BEV context, such as lane topology, intersection layout, and long-range spatial relationships.

Increasing the number to 128 causes a larger degradation, with $S_{\text{num}}$ dropping by 7.5\% and $S_{\text{sem}}$ by 1.0\%. This indicates that excessive scene queries may weaken the bottleneck effect of the scene branch and introduce redundant or noisy spatial cues, causing the decoder to attend to less relevant regions and reducing the effectiveness of metric reasoning and semantic grounding.

\begin{figure*}[t]
\centering
\includegraphics[width=\linewidth]{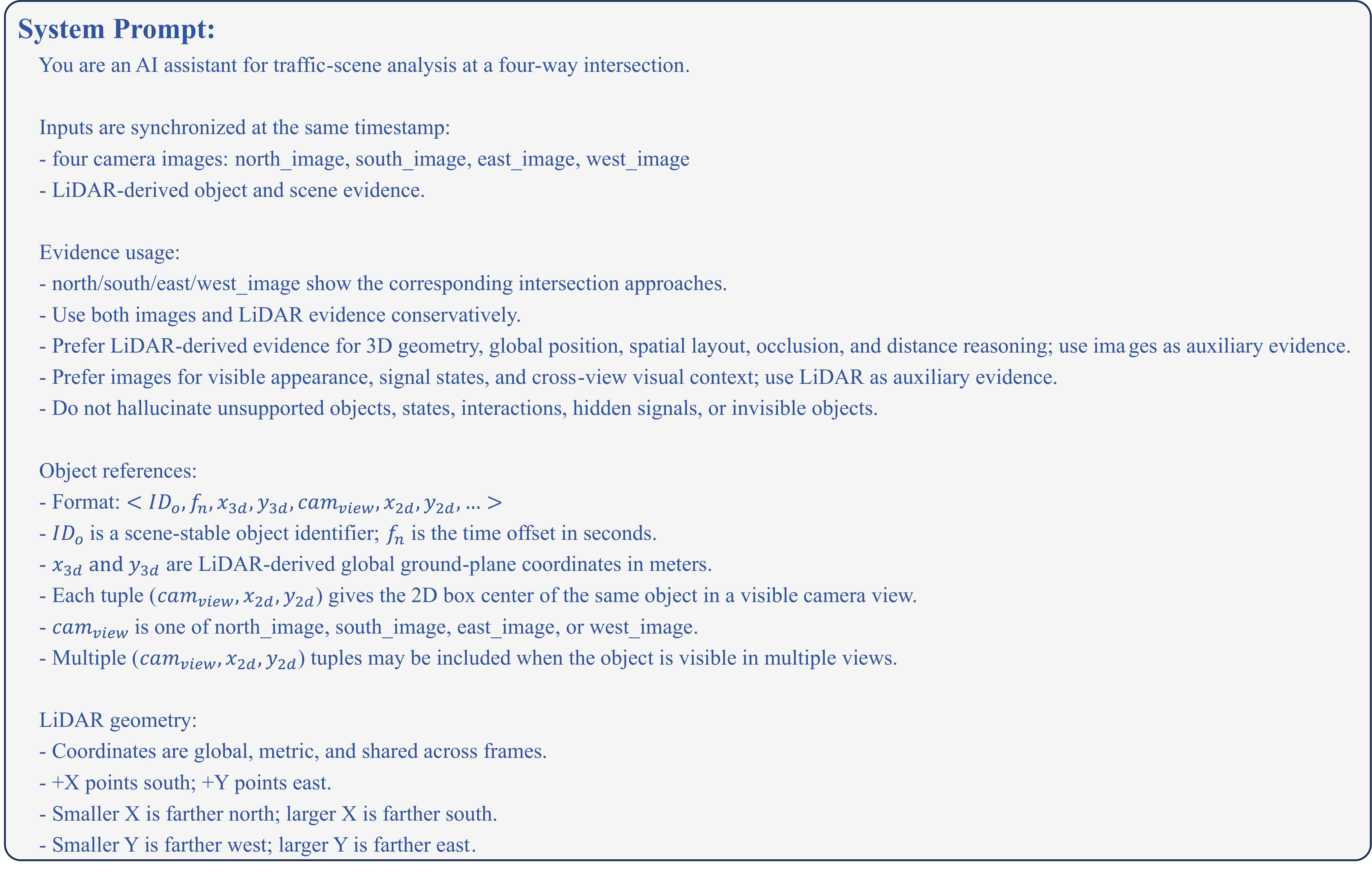}
\caption{System prompt for Inter-3D VQA evaluation. For image-based VLMs, LiDAR-related instructions are removed to avoid unavailable-modality interference.}
\label{fig:system_prompt}
\end{figure*}
\subsection{Numerical Error Analysis}
Table~\ref{tab:model_category_results} reports detailed numerical errors on grounded Inter-3D VQA across different numerical categories. Compared with image-based open-source, driving-specific, and roadside-specific VLMs, Inter-Geo achieves the best performance on most geometry-related metrics. In particular, it obtains the lowest errors on distance, 3D coordinates, and waypoint prediction, with errors of 0.15, 0.48, and 0.31, respectively. These results demonstrate the advantage of incorporating object- and scene-level aligned LiDAR representations, which provide explicit geometric cues for spatial estimation, 3D localization, and trajectory-related reasoning. This is especially beneficial for tasks where vision-only models often suffer from depth ambiguity and limited metric awareness.

However, Inter-Geo does not outperform all baselines across every category. For 2D coordinate errors, it remains competitive but is not clearly superior, which is expected because this metric depends more on image-plane localization than 3D spatial grounding. In addition, its acceleration error remains relatively high compared with several baselines, indicating that fine-grained dynamic attributes are still challenging. Since acceleration is sensitive to temporal estimation and can be noisy, it is difficult to infer reliably from limited LiDAR representations and short-term observations. These limitations suggest future improvements in tighter image--LiDAR alignment and stronger temporal motion modeling.

\subsection{System Prompt}
We design a dedicated system prompt for Inter-3D VQA to ensure consistent multimodal reasoning during evaluation. As shown in Fig.~\ref{fig:system_prompt}, the prompt specifies the available inputs, including four directional camera views and LiDAR-derived object and scene evidence. It further defines evidence usage rules: LiDAR evidence is prioritized for 3D geometry, global position, spatial layout, occlusion, and distance reasoning, while images are primarily used for visible appearance, signal states, and cross-view visual context. Note that the LiDAR evidence described in the prompt is used for grounding guidance. In Inter-Geo, the actual LiDAR information is provided as external feature tensors, aligned by DQNet, and injected into the decoder through LDA.

For image-based VLMs, we remove all LiDAR-related instructions from the prompt, including LiDAR evidence descriptions, coordinate-system definitions, and point-cloud-based geometric rules. This prevents unavailable modality information from interfering with image-only models, while keeping the image inputs, object references, and general reasoning constraints consistent across models.

To support grounded reasoning, the prompt also includes a unified object reference format containing scene-stable object IDs, temporal offsets, global coordinates, and multi-view 2D projections. In addition, it specifies the coordinate system and geometric rules for distance, lane, crosswalk, zone, and relative-direction reasoning when such information is available. We apply modality-consistent prompts across open-source, closed-source, and Inter-Geo models to ensure fair and controlled evaluation.

\section{Dataset Examples}
\subsection{Question Templates}
\begin{figure*}[t]
\centering
\includegraphics[width=\linewidth]{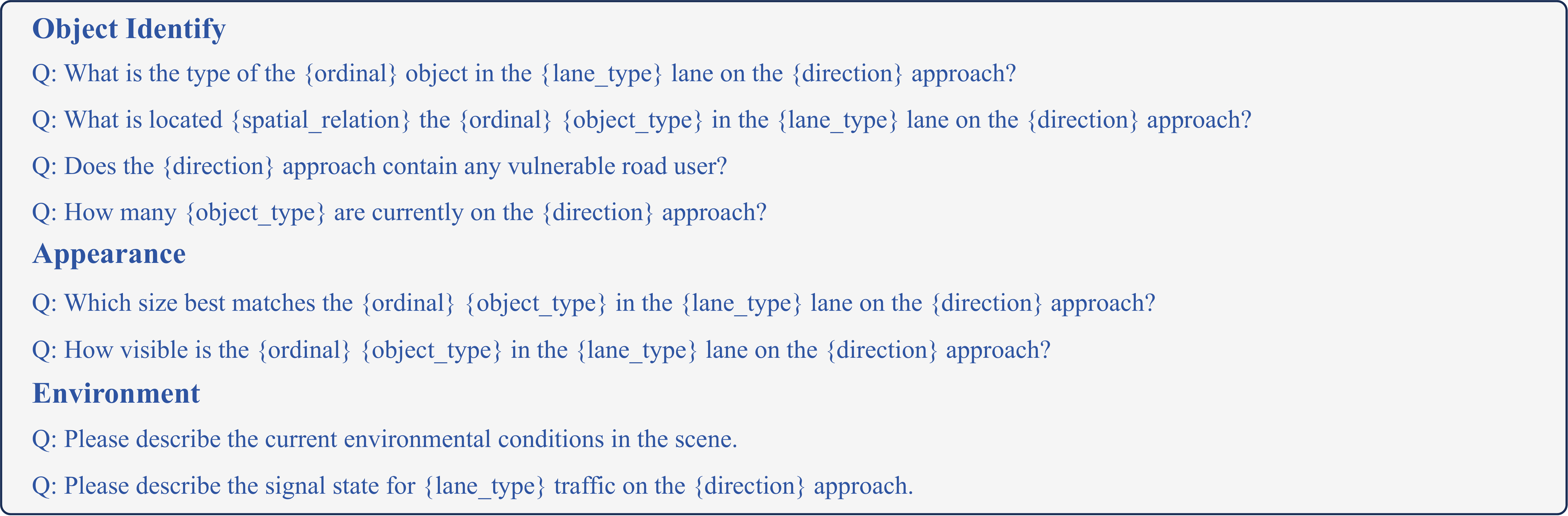}
\caption{Representative free-form question templates for Basic Perception.}
\label{fig:basic_perception_free}
\end{figure*}

\begin{figure*}[t]
\centering
\includegraphics[width=\linewidth]{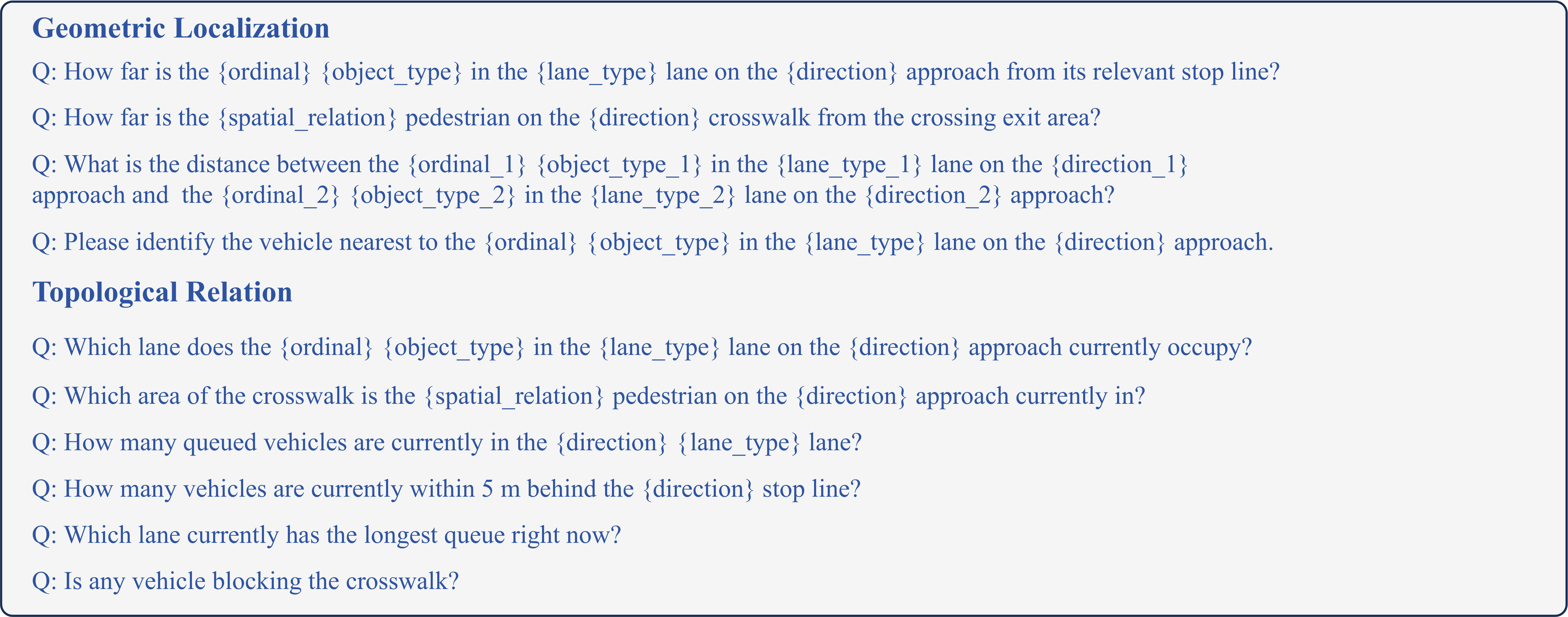}
\caption{Representative free-form question templates for Spatial Reasoning.}
\label{fig:spatial_reasoning_free}
\end{figure*}

\begin{figure*}[t]
\centering
\includegraphics[width=\linewidth]{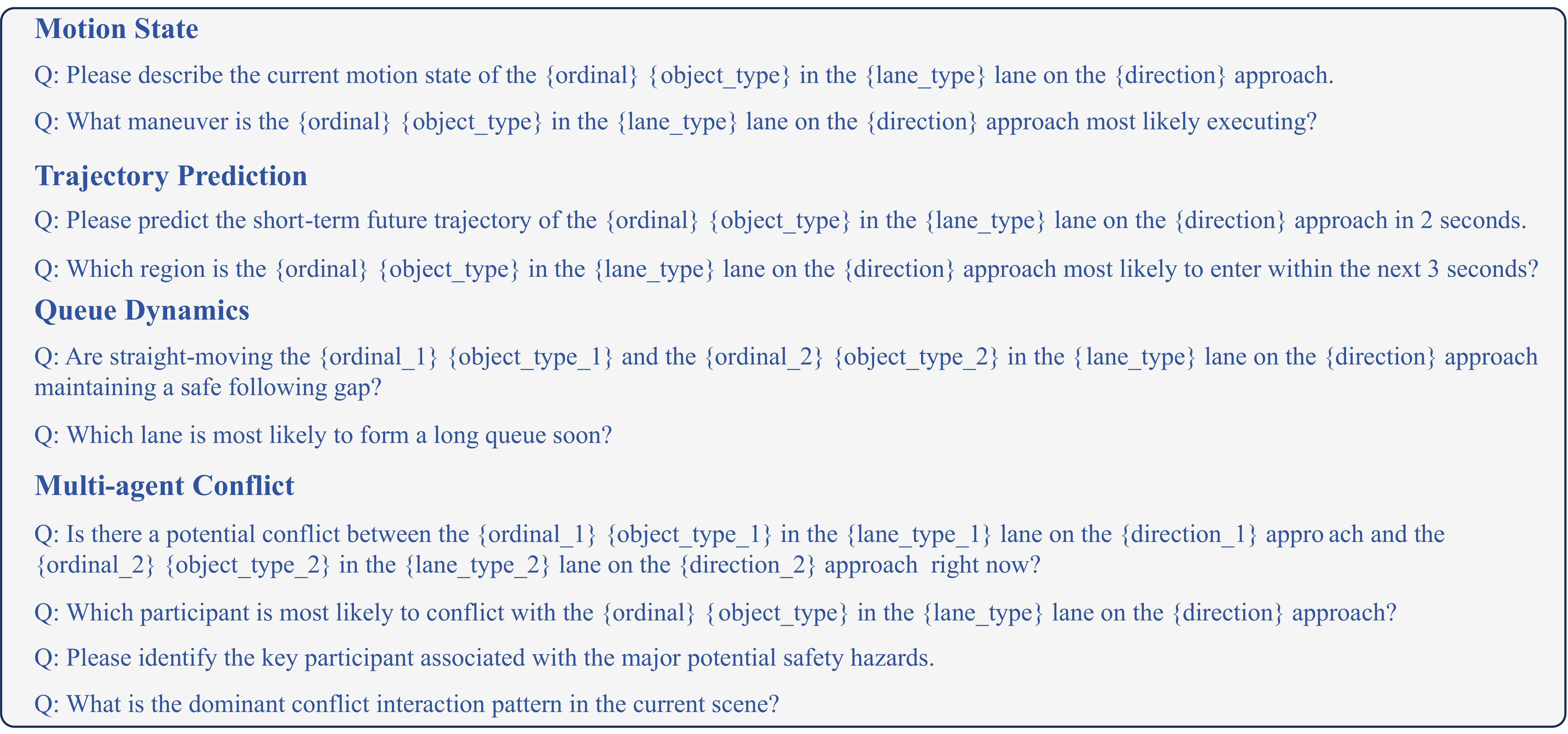}
\caption{Representative free-form question templates for Temporal Reasoning.}
\label{fig:temporal_reasoning_free}
\end{figure*}

\begin{figure*}[t]
\centering
\includegraphics[width=\linewidth]{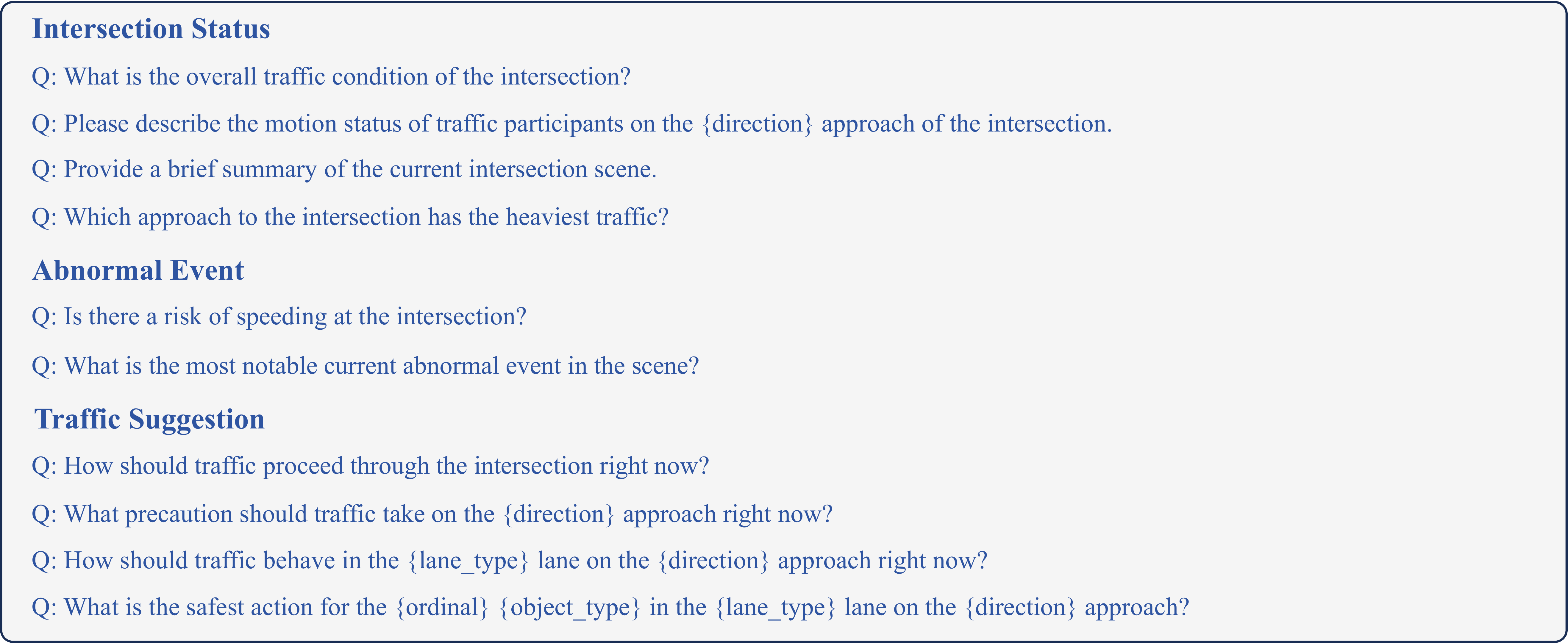}
\caption{Representative free-form question templates for Scene Understanding.}
\label{fig:scene_understanding_free}
\end{figure*}

\begin{figure*}[t]
\centering
\includegraphics[width=\linewidth]{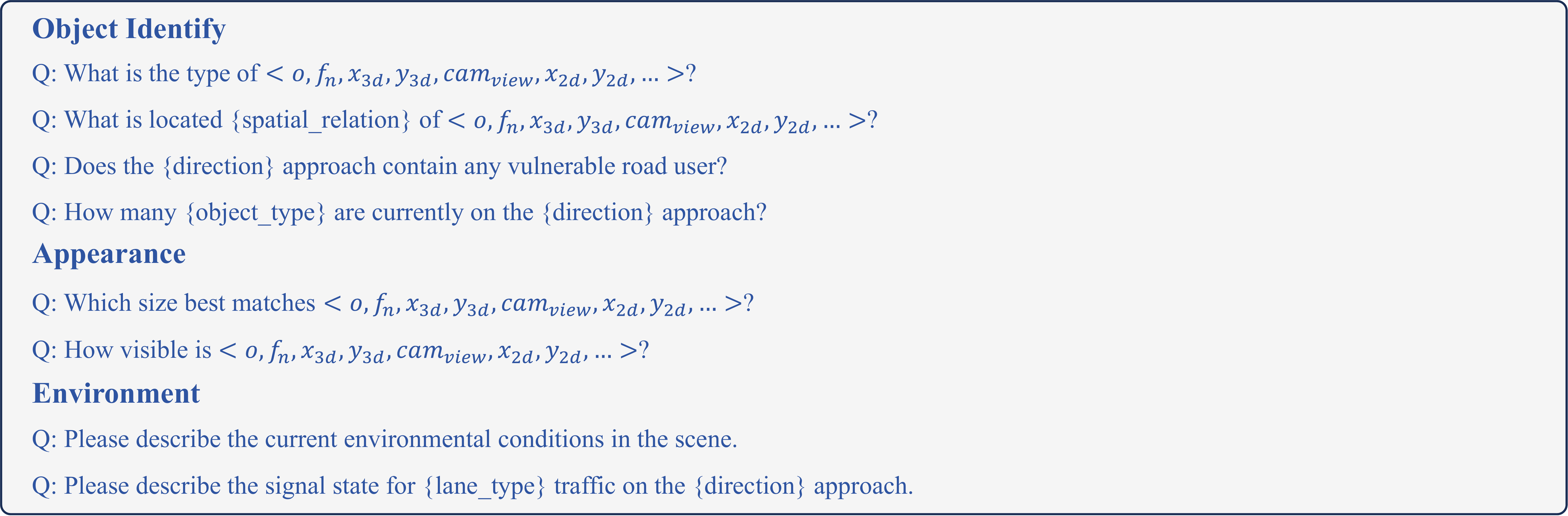}
\caption{Representative grounded question templates for Basic Perception.}
\label{fig:basic_perception_ground}
\end{figure*}

\begin{figure*}[t]
\centering
\includegraphics[width=\linewidth]{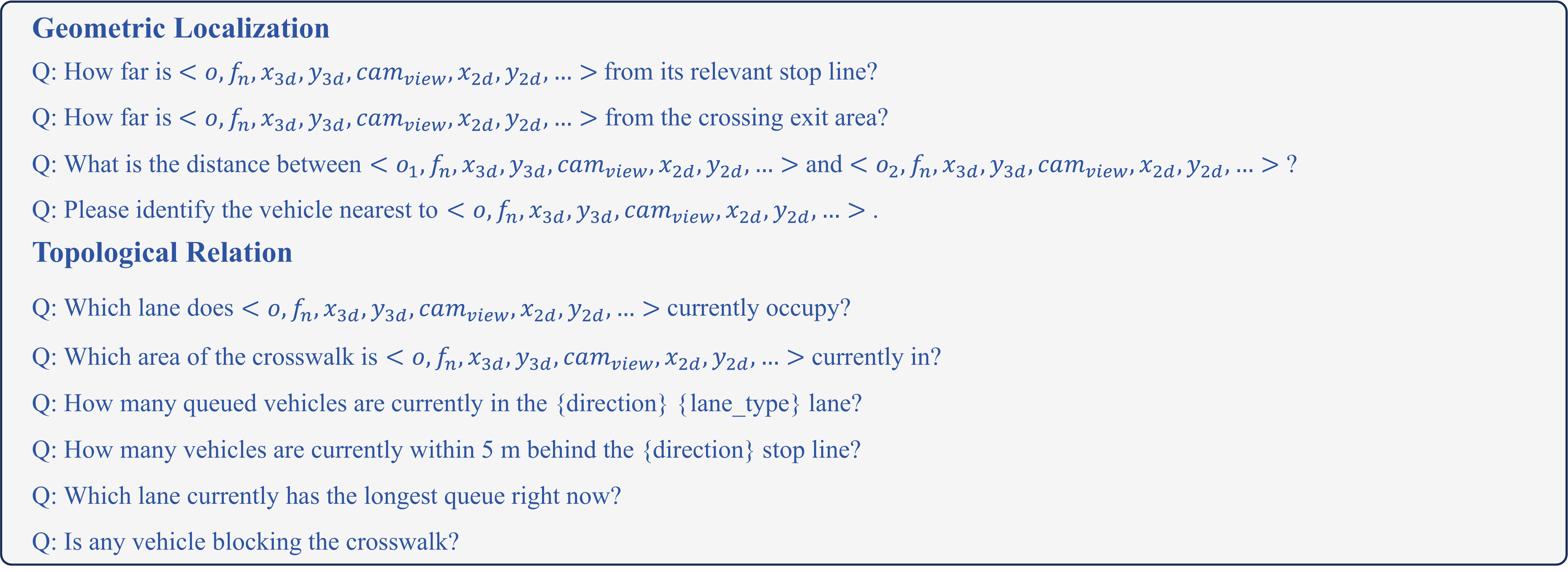}
\caption{Representative grounded question templates for Spatial Reasoning.}
\label{fig:spatial_reasoning_ground}
\end{figure*}

\begin{figure*}[t]
\centering
\includegraphics[width=\linewidth]{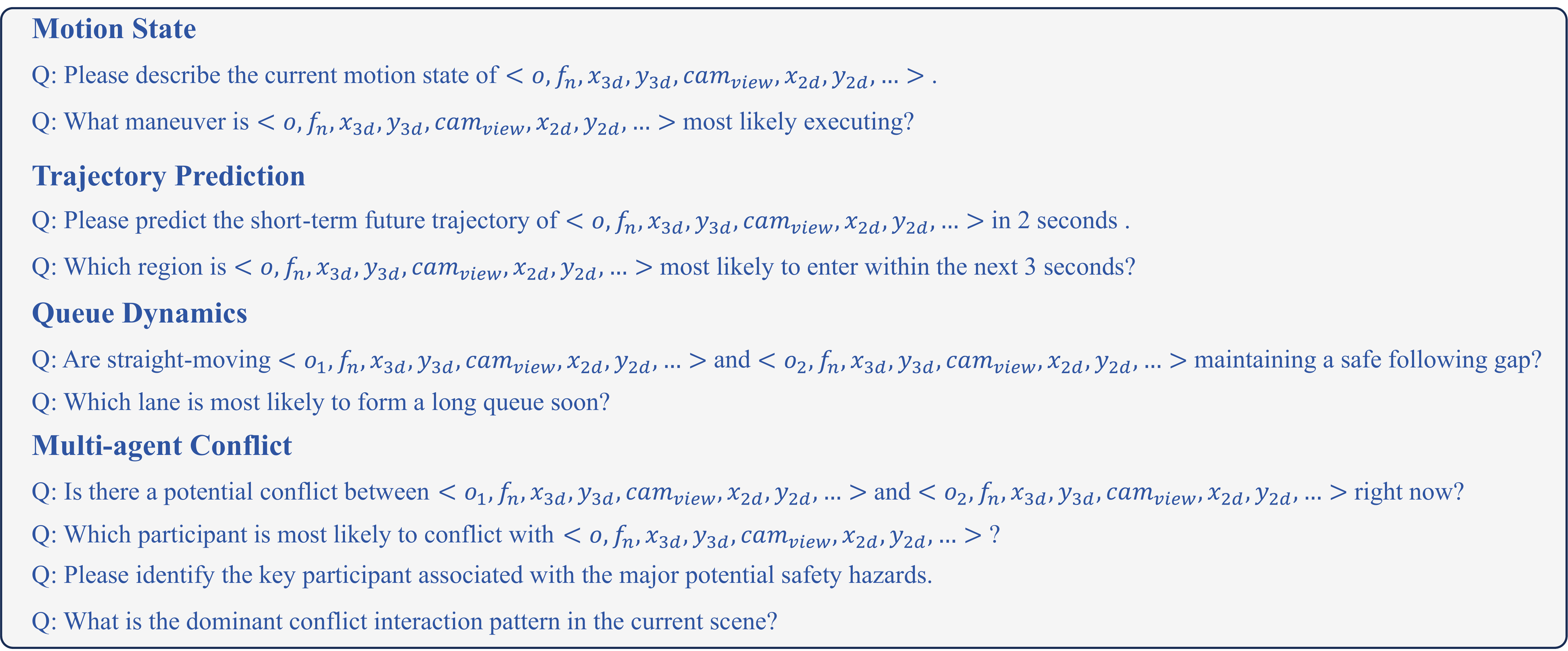}
\caption{Representative grounded question templates for Temporal Reasoning.}
\label{fig:temporal_reasoning_ground}
\end{figure*}

\begin{figure*}[t]
\centering
\includegraphics[width=\linewidth]{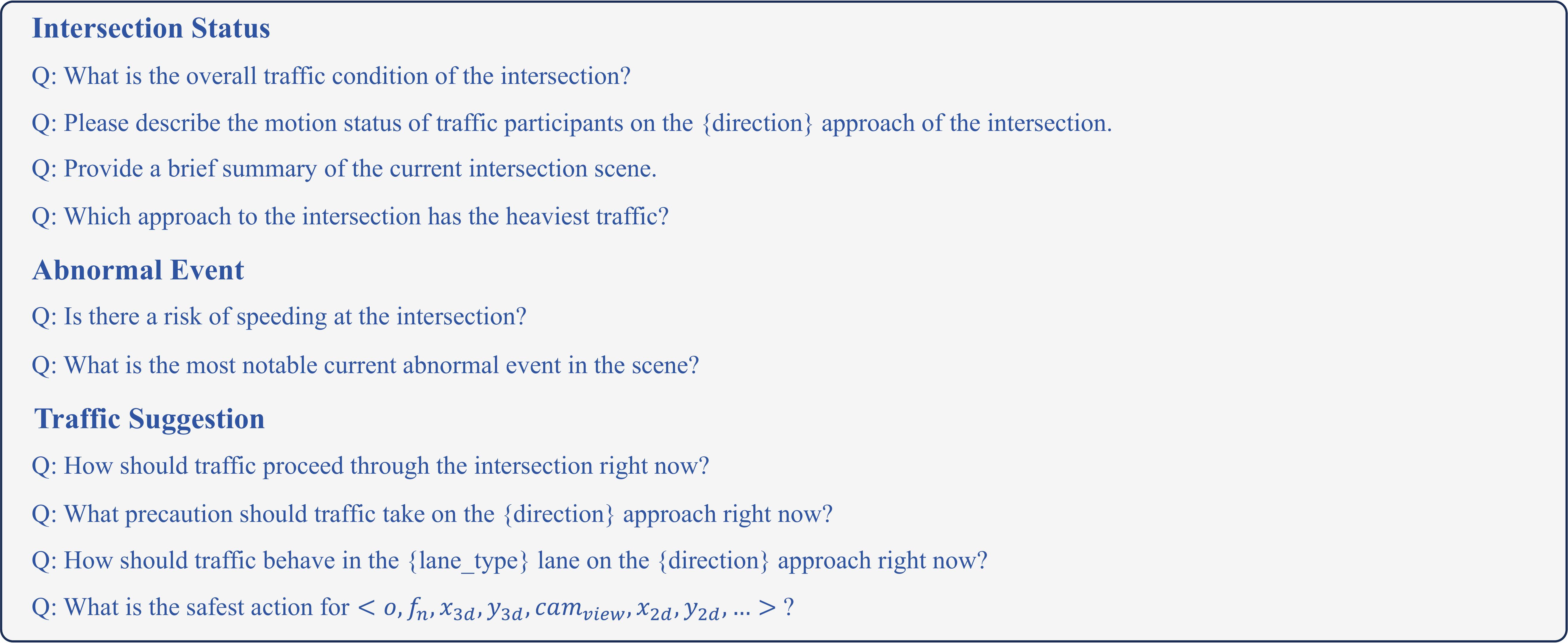}
\caption{Representative grounded question templates for Scene Understanding.}
\label{fig:scene_understanding_ground}
\end{figure*}

In this section, we provide representative examples of question templates for each task category. Inter-3D VQA contains two complementary QA formats: free-form QA and grounded QA. Figs.~\ref{fig:basic_perception_free}--\ref{fig:scene_understanding_free} illustrate representative free-form templates, while the corresponding grounded templates are shown in Figs.~\ref{fig:basic_perception_ground}--\ref{fig:scene_understanding_ground}. Both formats cover the four task categories: Basic Perception, Spatial Reasoning, Temporal Reasoning, and Scene Understanding.

Within each category, templates are further organized into multiple subcategories, each targeting specific reasoning capabilities, such as object identification, geometric measurement, motion analysis, and interaction understanding. This design enables systematic coverage of diverse reasoning patterns in complex intersection scenarios.

During dataset construction, these placeholders are automatically instantiated using structured scene metadata extracted from synchronized multi-view images, LiDAR point clouds, object annotations, and HD maps. Specifically, \texttt{\{ordinal\}} represents object order in the same lane (e.g., first, second), \texttt{\{object\_type\}} denotes object category (e.g., car, truck), \texttt{\{lane\_type\}} indicates lane type (e.g., left-turn, right-turn), \texttt{\{direction\}} specifies approach direction (e.g., north, east), and \texttt{\{spatial\_relation\}} encodes relative spatial relations (e.g., in front of, on the right, nearest). This process enables scalable and controllable QA generation while preserving semantic consistency across different reasoning tasks. By explicitly incorporating infrastructure topology and spatial relationships into template construction, Inter-3D VQA supports fine-grained multimodal reasoning beyond conventional image-centric VQA benchmarks.

\subsection{Qualitative Evaluations}
\begin{figure*}[t]
\centering
\includegraphics[width=\linewidth]{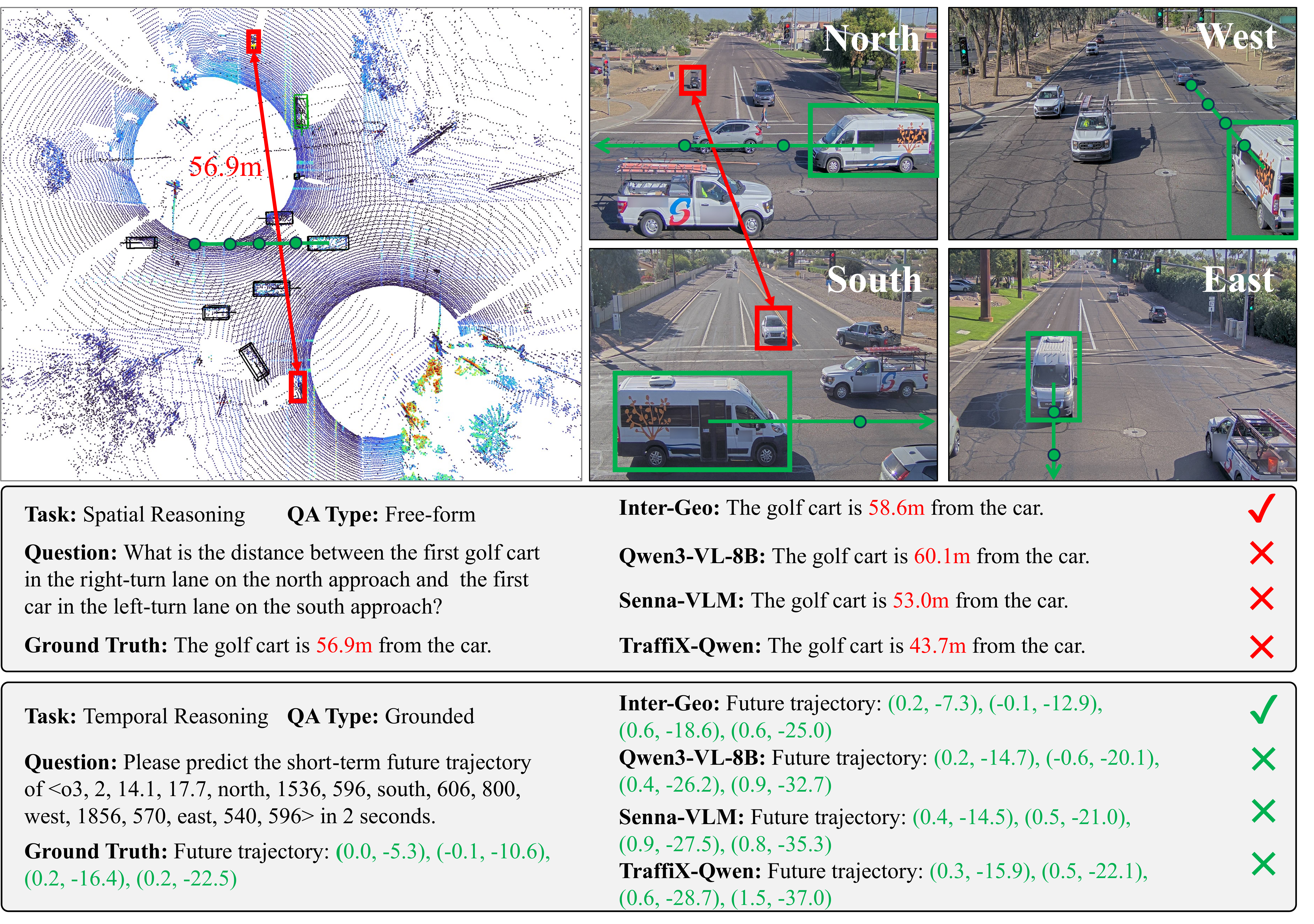}
\caption{Daytime examples of Spatial Reasoning and Temporal Reasoning.}
\label{fig:example1}
\end{figure*}

\begin{figure*}[t]
\centering
\includegraphics[width=\linewidth]{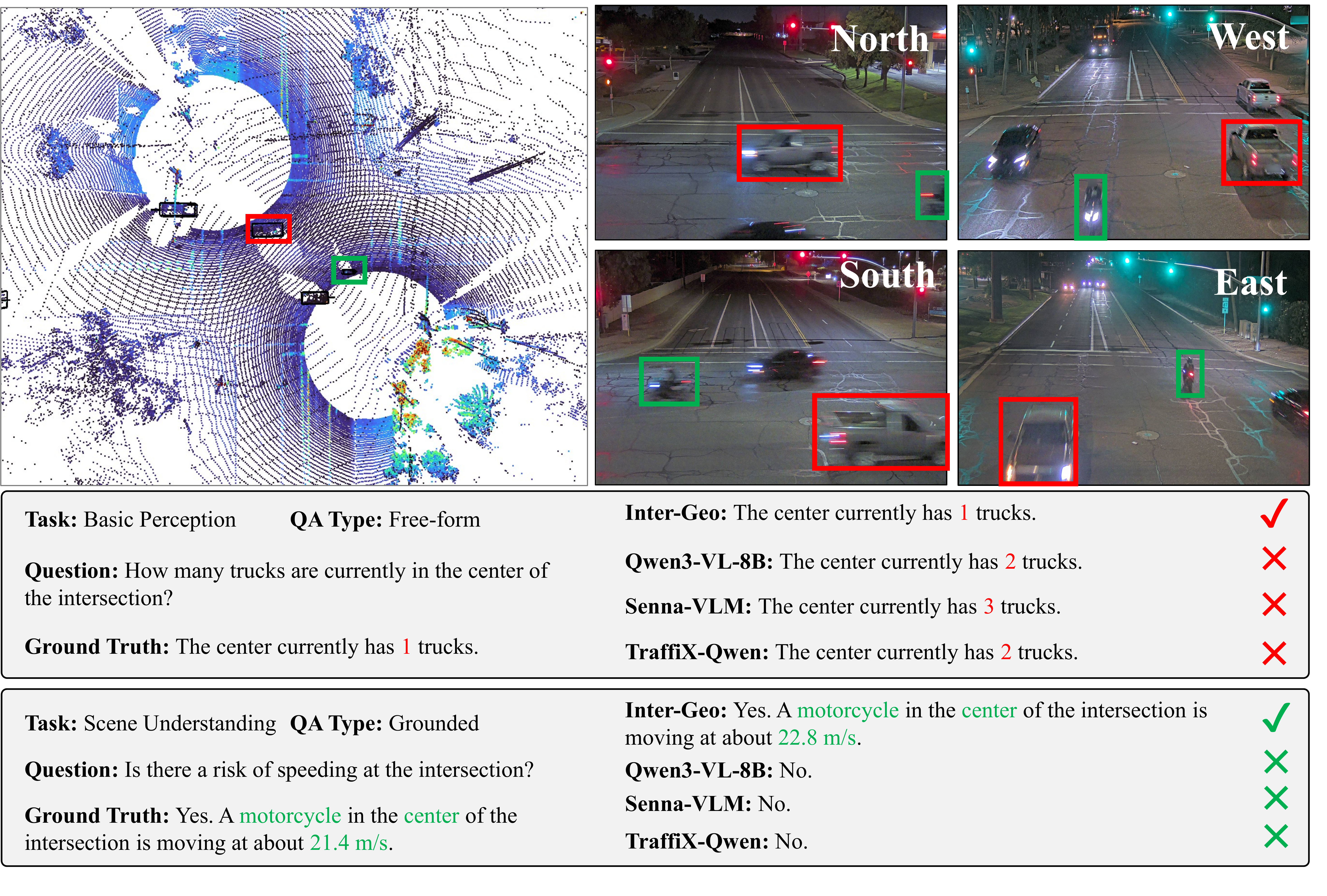}
\caption{Nighttime examples of Basic Perception and Scene Understanding.}
\label{fig:example2}
\end{figure*}

We visualize representative examples from the test set in Figs.~\ref{fig:example1} and~\ref{fig:example2}. Each example includes the global point cloud, multi-view traffic images, the question, ground-truth answer, and predictions from Inter-Geo and several representative baseline models.

Figure~\ref{fig:example1} illustrates spatial and temporal reasoning cases in a dense daytime traffic scenario. In the spatial reasoning example, the target golf cart and car appear in different camera views, requiring cross-view association and distance estimation in the global 3D point cloud. Inter-Geo predicts a distance of 58.6 m, close to the ground truth of 56.9 m, while image-based baselines show larger errors. This demonstrates the benefit of explicit LiDAR geometry for distance estimation. The temporal reasoning example requires predicting the two-second future trajectory of a moving van. Inter-Geo better preserves the trajectory direction and displacement trend with only minor deviation from the ground truth, whereas other models show larger longitudinal or lateral drift. This suggests that aligned LiDAR representations provide effective geometric cues for trajectory-level spatial reasoning.

Figure~\ref{fig:example2} illustrates basic perception and scene understanding in a fast-moving nighttime scenario. In the basic perception case, nighttime illumination, motion blur, and occlusion make object counting challenging. Inter-Geo correctly identifies one truck in the intersection center, while other models overestimate the count due to misidentification of surrounding vehicles. In the scene understanding case, Inter-Geo detects the speeding-risk event and provides the relevant object category and speed estimate, whereas the baseline models fail to recognize the potential hazard. These examples show that global point clouds improve localization, motion estimation, and safety-critical reasoning.

\end{document}